%% file: main.tex
\pdfoutput=1
\documentclass{article}
\input{preamble}

\title{\centering \Large Causal Memory Policy: Making Memory Utility Identifiable by Intervening on Retrieval}

\author{
  Arman Behnam\\
  Department of Computer Science\\
  Illinois Institute of Technology\\
   Chicago, IL\\
  \texttt{abehnam@hawk.illinoistech.edu} \\
  \And
  Binghui Wang \\
  Department of Computer Science\\
  Illinois Institute of Technology\\
   Chicago, IL\\
  \texttt{bwang70@illinoistech.edu} \\
}

\begin{document}

\maketitle
\lhead{Preprint}
\vskip 0.3in

\input{sections/0_abstract}
\input{sections/1_introduction}
\input{sections/2_problem}
\input{sections/3_method}
\input{sections/4_experiments}
\input{sections/5_conclusion}
\input{sections/6_statements}
\newpage

\bibliographystyle{plainnat}
\bibliography{ref}

\clearpage
\part*{Appendix}
\appendix
\label{sec:appendix}

\startcontents[appendix]
\printcontents[appendix]{}{1}{\setcounter{tocdepth}{2}}

\clearpage

\input{appendix/1_related_work}
\input{appendix/2_theory}
\input{appendix/3_experiments}
\input{appendix/4_limitations}
\input{appendix/5_applications}
\end{document}

%% file: preamble.tex
\PassOptionsToPackage{numbers, compress}{natbib}

\usepackage{iclr2027_conference,times}
\iclrfinalcopy

\usepackage[utf8]{inputenc}
\usepackage[T1]{fontenc}
\usepackage{xurl}
\usepackage{hyperref}
\usepackage{booktabs}
\usepackage{amsfonts}
\usepackage{nicefrac}
\usepackage{microtype}
\usepackage[table]{xcolor}
\usepackage{graphicx}
\usepackage{amsmath, amssymb, amsthm}
\usepackage{mathtools}
\usepackage{enumitem}
\usepackage{pifont}
\usepackage{wrapfig}
\usepackage{diagbox}
\usepackage{multirow}
\usepackage{algorithm}
\usepackage{algorithmic}
\usepackage{slashed}
\usepackage{braket}
\usepackage{orcidlink}
\usepackage{bbm}
\usepackage{hhline}
\usepackage{subcaption}
\usepackage{soul}
\usepackage{float}
\usepackage{thmtools,thm-restate}
\usepackage{tikz}
\usetikzlibrary{positioning, fit, backgrounds, calc, arrows.meta}
\pgfdeclarelayer{background}
\pgfsetlayers{background,main}
\usepackage{bm}
\usepackage{mathrsfs}
\usepackage{titletoc}

\definecolor{myred}{RGB}{215,48,39}
\definecolor{mygreen}{RGB}{26,152,80}

\newtheorem{proposition}{Proposition}
\newtheorem{corollary}{Corollary}
\newtheorem{definition}{Definition}
\newtheorem{assumption}{Assumption}
\newtheorem{remark}{Remark}
\newtheorem{comparatorcond}{Condition}
\newcommand{\E}{\mathbb{E}}

\newcommand{\ind}{\mathbf{1}_}
\newcommand{\doo}[1]{\text{do}(#1)}

\newcommand{\cM}{\mathcal{M}}
\newcommand{\cU}{\mathcal{U}}
\newcommand{\cV}{\mathcal{V}}
\newcommand{\cF}{\mathcal{F}}
\newcommand{\cQ}{\mathcal{Q}}
\newcommand{\cO}{\mathcal{O}}

\newcommand{\cN}{\mathcal{N}}
\newcommand{\cT}{\mathcal{T}}

\newcommand{\cP}{\mathcal{P}}
\newcommand{\cL}{\mathcal{L}}

\newif\ifshowchanges
\showchangesfalse

\renewcommand{\algorithmicrequire}{\textbf{Input:}}
\renewcommand{\algorithmicensure}{\textbf{Output:}}

\definecolor{linkblue}{HTML}{2C5F8A}
\hypersetup{colorlinks=true, linkcolor=linkblue, citecolor=linkblue, urlcolor=linkblue,
  pdftitle={Causal Memory Policy: Making Memory Utility Identifiable by Intervening on Retrieval},
  pdfauthor={Arman Behnam, Binghui Wang}}

%% file: sections/0_abstract.tex
\begin{abstract}
Memory-augmented large language models must decide which memories to retain, and recent systems do so by estimating each memory's effect on task performance. However, these estimates rely entirely on retrieved memories. When a memory is never retrieved, store-level interventions produce identical outcomes, leaving its utility unidentified. This is a retrieval-level positivity violation, invisible to diagnostics that examine only memory operations.

We introduce Causal Memory Policy (CMP), a causal framework that restores identification by intervening on retrieval itself, reserving a fixed number of context slots for memories sampled with known propensity scores. CMP estimates memory utility via self-normalized inverse propensity weighting under a balanced assignment design with fixed sampling propensities. We prove the causal factorization of memory utility through retrieval, the unbiasedness and exact variance of the estimator, and the optimal decision rule under irreversible operations. Empirically, identification fails for $54\%$ of required memories on LongMemEval and $67\%$ on LoCoMo, and the failure persists in a deployed memory system. CMP improves discrimination between required and non-required memories from $0.54$ to $0.66$ AUC. Finally, we show that identified memory utility alone is insufficient for retention decisions: per-query utility reaches $0.78$ AUC on the query for which it is estimated. {Code is available at: \url{https://anonymous.4open.science/r/cmp-release-D0C3/}}.
\end{abstract}

%% file: sections/1_introduction.tex
\section{Introduction}
\label{sec:introduction}
\vspace{-2mm}
Large language models (LLMs) exhibit remarkable capabilities across a broad spectrum of tasks, but remain fundamentally constrained by fixed context windows that limit the information accessible at inference time.
As the scope and duration of LLM deployments grow, effective management of what is remembered, forgotten, and reorganized becomes a first-order design challenge. Memory-augmented LLMs~\citep{zhong2024memorybank, wang2024survey} address this limitation by maintaining external memory stores that persist across interactions, enabling long-horizon reasoning, user personalization, and incremental knowledge accumulation. {A detailed discussion of related work is in Appendix~\ref{app:related}.}

Deciding what to retain requires estimating the utility of each memory. Recent systems estimate this quantity from per-memory success and failure counts~\citep{simsek2026forget}, paired trajectories with and without a candidate memory~\citep{gao2026skillaudit}, controlled interventions on the candidate set~\citep{srivastava2026causal}, and explicit retention budgets~\citep{wu2026forget}. All of these estimators rely solely on memories already retrieved by the logging policy (see Appendix~\ref{app:examples} for an example). This raises four fundamental questions:
(1)~\emph{When is memory utility identified from logged interactions?} Identification requires observing a memory both present and absent in the retrieved context, which no deterministic logging policy guarantees.
(2)~\emph{What does an estimator return when the estimand is not identified?} A degenerate contrast often yields an estimate near zero, making identification failure indistinguishable from genuinely low memory utility.
(3)~\emph{Which memory management policies are affected?} Only policies that rank memories by estimated utility are susceptible; recency- or frequency-based heuristics never evaluate the affected memories.
(4)~\emph{How can identification be restored, and at what cost?} Randomization must be introduced into retrieval, since perturbing the memory store leaves the retrieved context unchanged whenever the perturbed memory is not retrievable.

The root cause is that \textbf{retrieval mediates the effect of the memory state on task output}. A memory influences an answer only if it is retrieved into the context the model reads, so intervening on the memory store affects the output only through what retrieval subsequently surfaces. Memory utility therefore factorizes into two components: the probability of retrieval and the contribution once retrieved. Logged interactions cannot distinguish a memory with genuinely low utility from one that was never given the opportunity to contribute. This is a \emph{positivity violation}, the causal inference condition requiring every unit to have a nonzero probability of receiving each treatment. Because the violation occurs at the level of retrieval, diagnostics that verify exploration only over memory operations fail to detect it. 
Existing causal approaches to memory management do not address this retrieval-level positivity failure. They either place the ground-truth answer inside the selection loop~\citep{srivastava2026causal}, which is unavailable at deployment, or verify overlap only over the operations selected by the policy~\citep{li2026mempo}, leaving retrieval itself uncontrolled.

We propose a corrected view: {\bf memory utility as a retrieval-mediated causal effect}. We formalize a memory-augmented LLM as a structural causal model (SCM)~\citep{pearl2009causality}, where retrieval mediates the effect of memory operations on task outputs. Structural causal models have also been used to explain graph neural network predictions through their causal subgraphs~\citep{behnam2024graph}, to learn representations when labels cause the observed features~\citep{behnam2026measure}, and to learn causal representations without a known causal structure~\citep{behnam2026reasoning}; here they expose a support condition that associational memory scores leave unchecked. Built on this formulation, we develop the {\bf Causal Memory Policy (CMP)}, whose components resolve the above four questions:
(i)~{\bf Identification}. We derive a factorization of memory utility into retrievability and conditional usefulness, prove the resulting non-identifiability under deterministic retrieval, and explain why unidentified estimators converge to values near zero rather than becoming undefined, answering Questions~(1) and~(2).
(ii)~{\bf Restoring identification and cost}. We introduce a randomized exposure design that reserves $k$ context slots for memories sampled with known propensity, restoring positivity by construction while requiring a budget that scales only with the candidate pool being identified, answering Question~(4).
(iii)~{\bf Empirical characterization}. We quantify retrieval-level positivity failure on two benchmarks and a deployed memory system, identifying when utility-based policies are affected and when heuristic policies are not, answering Question~(3).
(iv)~{\bf Decision theory}. We characterize what identified memory utility can support by deriving the Bayes-optimal policy under asymmetric reversible and irreversible operations, and show that no aggregation available to a retention policy predicts a memory's utility on unseen queries.

We evaluate CMP on LongMemEval~\citep{wu2024longmemeval} (long-horizon retention over $\sim\!490$-turn histories), LoCoMo~\citep{maharana2024evaluating} (multi-session dialogue with many queries over a single memory store), HotpotQA~\citep{yang2018hotpotqa} and MuSiQue~\citep{trivedi2022musique} (2--4-hop reasoning, as replication), and Mem0~\citep{chhikara2025mem0} (a deployed memory system, as transfer). \emph{We find that retrieval-level identification failure is substantial rather than an implementation artifact: it affects $54\%$ of required memories on LongMemEval and $67\%$ on LoCoMo, and persists in Mem0, where unretrieved evidence has a median retrieval rank of $139$ among $291$ stored memories.} Store-level randomization leaves the retrieved context unchanged on $52.2\%$ of perturbations, causing utility-based eviction to perform no better than chance (lift $0.97$). Restricting evaluation to retrieved memories eliminates the gap between the two estimators ($0.655$ vs.\ $0.658$ AUC), localizing the failure to positivity. CMP raises discrimination between required and non-required memories from $0.54$ to $0.66$ AUC, and per-query estimation raises it to $0.78$. Finally, we show that identified utility alone does not determine retention: no aggregation over observed queries predicts a memory's value on unseen queries, indicating that the remaining limitation lies in the estimand.

%% file: sections/2_problem.tex
\vspace{-5mm}
\section{Problem Formulation}
\label{sec:problem}

\vspace{-3mm}
We formalize a memory-augmented LLM as an SCM and memory operations as causal interventions. We show that retrieval mediates all paths from memory state to task output and derive the identification condition imposed by this mediation. A summary of the notation is in Table~\ref{tab:notation} in the Appendix \ref{app:notation}.

\vspace{-3mm}
\subsection{Memory System as a Structural Causal Model}
\label{sec:scm}
\vspace{-2mm}
\begin{definition}[Memory-Augmented LLM as SCM]
\label{def:scm}
A memory-augmented LLM is an SCM $\cM = (\cU, \cV, \cF)$. Exogenous variables $\cU = \{Q, C\}$ consist of user queries $Q$ drawn from a task distribution $\cQ$ and observed context $C$ (conversation metadata, task type), with realizations $q_t$ and $c_t$.
Endogenous variables $\cV = \{M, R, O, Y\}$ are defined as: $M_t \in 2^{\mathbb{M}}$ is the memory state at time $t$ and a finite set of stored memories from universe $\mathbb{M}$, $R_t \subseteq M_t$ is the retrieved subset for query $q_t \in Q$, $O_t \in \cO \triangleq \{\texttt{store}, \texttt{forget}, \texttt{consolidate}, \texttt{noop}\}$ {is the memory operation applied at time $t$, each parameterized by its arguments as in Definition~\ref{def:intervention}}, and $Y_t \in \mathcal{Y}$ is the output of a downstream task. The structural equations $\cF$ are $R_t = f_R(M_t, q_t)$ (retrieval), $Y_t = f_Y(R_t, q_t, \theta_{\text{LLM}})$ (generation), $O_t = f_O(M_t, q_t, c_t)$ (operation selection), and $M_{t+1} = f_M(M_t, O_t, q_t)$ (memory transition).
\end{definition}
\vspace{-2mm}
Here, $c_t$ denotes the observed context (e.g., conversation metadata and task type) on which the policy conditions when selecting an operation, and $\theta_{\text{LLM}}$ denotes the frozen LLM parameters. Note that $c_t$ enters $f_O$ but not $f_Y$: the generator receives only the retrieved memories and the query, so every path from the memory state to the output passes through $R_t$. This mediation underlies the identification result in Section~\ref{sec:identification} and holds by construction in all our experiments: the generator takes as input only the retrieved memory texts and the question, without any item metadata.

\begin{wrapfigure}{r}{0.75\textwidth}
\vspace{-4mm}
    \centering
     \scalebox{0.68}{\begin{tikzpicture}[node distance=1.3cm and 1.6cm,
        main/.style={draw, circle, minimum size=1.0cm, font=\small, thick},
        interv/.style={draw, circle, minimum size=1.0cm, font=\small, thick, fill=blue!12},
        shadow/.style={draw, circle, minimum size=1.0cm, font=\small, thick, densely dashed, fill=gray!12, text=gray!55},
        conf/.style={draw=none, rectangle, font=\small\itshape, minimum size=0pt},
        arr/.style={->, >=stealth, thick},
        confarr/.style={->, >=stealth, thick, densely dashed, color=red!55},
        intarr/.style={->, >=stealth, thick, color=blue!65},
        medarr/.style={->, >=stealth, very thick, color=blue!70},   % [NEW]
        shadarr/.style={->, >=stealth, densely dashed, color=gray!50},
        chalannot/.style={draw=none, font=\scriptsize\itshape, text width=4cm, align=center, color=#1},
        timebox/.style={draw, thick, rounded corners=4pt, inner sep=10pt, densely dotted, color=gray!45},
        expo/.style={draw, circle, minimum size=0.85cm, font=\small, thick, fill=green!15, densely dashed},
        expoarr/.style={->, >=stealth, very thick, color=green!50!black},]
        \node[main] (Mt) {$M_t$};
        \node[main, right=1.5cm of Mt] (Rt) {$R_t$};
        \node[main, right=1.5cm of Rt] (Yt) {$Y_t$};
        \node[interv, below=1.5cm of Rt] (Ot) {$O_t$};
        % \node[conf, above=1.0cm of Rt] (Qt) {$q_t, c_t$};
        \node[conf, above=1.0cm of Rt] (Qt) {$q_t$};
        \node[conf, left=0.9cm of Ot] (Ct) {$c_t$};
        
        \draw[medarr] (Mt) -- (Rt);   % [CHANGED] was arr
        \draw[medarr] (Rt) -- (Yt);
        \draw[arr] (Mt) -- (Ot);
        \draw[confarr] (Qt) -- (Rt);
        \draw[confarr] (Qt) -- (Yt);
        \draw[confarr] (Qt) -- (Ot);
        \draw[confarr] (Ct) -- (Ot);
        
        \node[main, right=1.5cm of Yt, fill=blue!6] (Mt1) {$M_{t\!+\!1}$};
        \node[main, right=1.5cm of Mt1] (Rt1) {$R_{t\!+\!1}$};
        \node[main, right=2cm of Rt1] (Yt1) {$Y_{t\!+\!1}$};
        \node[interv, below=1.5cm of Rt1] (Ot1) {$O_{t\!+\!1}$};
        % \node[conf, above=1.0cm of Rt1] (Qt1) {$q_{t\!+\!1}, c_{t\!+\!1}$};
        \node[conf, above=1.0cm of Rt1] (Qt1) {$q_{t\!+\!1}$};
        \node[conf, left=0.9cm of Ot1] (Ct1) {$c_{t\!+\!1}$};

        \draw[medarr] (Mt1) -- (Rt1);   % [CHANGED] was arr
        \draw[medarr] (Rt1) -- (Yt1);
        \draw[arr] (Mt1) -- (Ot1);

        \draw[confarr] (Qt1) -- (Rt1);
        \draw[confarr] (Qt1) -- (Yt1);
        \draw[confarr] (Qt1) -- (Ot1);
        \draw[confarr] (Ct1) -- (Ot1);
        \draw[confarr] (Qt) to[bend left=28] (Mt1);

        \draw[intarr, thick] (Ot.east) -- node[below, font=\scriptsize, color=blue!65, yshift=-0.05cm] {$do(\cdot)$} (Mt1.south west);
        \draw[arr] (Mt) to[bend left=15] (Mt1);
        \draw[confarr] (Qt) to[bend left=28] (Mt1);
        
        \node[expo, below=0.9cm of Rt, xshift=1.6cm] (Zt) {$Z^{(t)}_m$};
        \draw[expoarr] (Zt) -- node[right, font=\scriptsize, color=green!45!black]
      {$\doo{m \in R_t}$} (Rt);
      
        \begin{scope}[on background layer]
            \node[timebox, 
            fit=(Mt)(Rt)(Yt)(Ot)(Qt)(Ct)(Zt),
                  label={[font=\small\bfseries, anchor=north west, xshift=2pt, yshift=-2pt]north west:Time $t$}] (tbox) {};
            \node[timebox, fit=(Mt1)(Rt1)(Yt1)(Ot1)(Qt1)(Ct1),
                  label={[font=\small\bfseries, anchor=north west, xshift=2pt, yshift=-2pt]north west:Time $t\!+\!1$}] (t1box) {};
        \end{scope}

        \node[chalannot=red!65, above=-0.1cm of Qt, xshift=0.0cm] (c1label) {\textbf{C1:} Query-induced confounding};
        \node[chalannot=blue!70, below=0.05cm of Mt, xshift=1.4cm, text width=4.2cm] (c2label) 
        {\textbf{C2:} Retrieval support \\ failure};
        \node[chalannot=orange!80!black, above=0.03cm of Mt1, xshift=0.8cm] (c3label)  {\textbf{C3:} Irreversible forgetting};

        \node[chalannot=purple!65, below=-0.1cm of Ot1, xshift=0.0cm] (c4label) 
        {\textbf{C4:} No oracle for optimal memory};

        \node[draw=none, font=\scriptsize, anchor=north west] at ([xshift=0.0cm, yshift=0.5cm]tbox.south west) (leg) {
        \begin{tabular}{@{}l@{\hspace{3pt}}l@{\hspace{10pt}}
                        l@{\hspace{3pt}}l@{}}
            \tikz\draw[->, >=stealth, thick] (0,0) -- (0.5,0); & Structural dependency &
            \tikz\draw[->, >=stealth, thick, densely dashed, color=red!55] (0,0) -- (0.5,0); & Query/context dependence \\
            \tikz\draw[->, >=stealth, very thick, color=blue!70] (0,0) -- (0.5,0); & Retrieval-mediated path &
            \tikz\draw[->, >=stealth, very thick, color=green!50!black] (0,0) -- (0.5,0); & Exposure intervention (CMP)
        \end{tabular}
        };
    \end{tikzpicture}}
    \vspace{-2mm}
    \caption{Causal DAG of memory-augmented LLMs across two timesteps.} 
    \label{fig:causal_graph}
    \vspace{-4mm}
\end{wrapfigure}

Figure~\ref{fig:causal_graph} presents the causal DAG of a memory-augmented LLM in two consecutive time steps. Black solid edges denote structural equations; red dashed edges denote confounding paths through the query $q_t$ and context $c_t$;  thick blue edges denote the mediating retrieval step. The DAG contains two sequential mediators with distinct roles. First, an operation $O_t$ affects future outputs only through the memory state it produces, giving the chain $O_t \rightarrow M_{t+1}$ with no direct path from an operation to a later output. This is why memory utility, rather than operation choice, is the estimand in Definition~\ref{def:utility}. Second, the memory state affects the output only through the retrieved subset, $M_t \rightarrow R_t \rightarrow Y_t$, since $Y_t$ depends on $M_t$ exclusively through $R_t=f_R(M_t,q_t)$. This second mediation determines whether the estimand is identifiable. Together, the two mediators form the causal chain $O_t \rightarrow M_{t+1} \rightarrow R_{t+1} \rightarrow Y_{t+1}$. The reverse edge $M_t \rightarrow O_t$ captures the policy selecting an operation based on current memory state. The graph remains acyclic as forward and reverse edges connect different time steps.

\vspace{-1mm}
Prior work states the first mediation and leaves the second implicit. \emph{Only the second governs identification}: an operation always induces a memory state, whereas a memory state containing $m$ need not retrieve $m$.  We therefore describe memory utility as \emph{retrieval-mediated} to emphasize that identifiability is determined by retrieval. Each memory operation corresponds to a concrete modification of the memory state. We formalize these as interventions in the SCM.
\begin{definition}[Memory Operations as Causal Interventions]
\label{def:intervention}
A memory operation $o \in \cO$ acts as an intervention $\doo{M_{t+1} = f_o(M_t)}$. $\texttt{store}(m_{\text{new}}, M_t)$ is $M_{t+1} = M_t \cup \{m_{\text{new}}\}$, $\texttt{forget}(m, M_t)$ is $M_{t+1} = M_t \setminus \{m\}$, $\texttt{consolidate}(m_i, m_j, M_t)$ is $M_{t+1} = (M_t \setminus \{m_i, m_j\}) \cup \{\textup{\textsc{Merge}}(m_i, m_j)\}$, where $\textup{\textsc{Merge}}(m_i, m_j)$ produces a single memory preserving content of both inputs $m_i$ and $m_j$, and $\texttt{noop}(M_t)$ is $M_{t+1} = M_t$. 
\vspace{-2mm}
\end{definition}

The central estimand is the causal effect of an individual memory on downstream task utility.
\begin{definition}[Memory Utility]
\label{def:utility}
The {utility} of memory $m$ at time $t$, given memory state $M_t$, is: 
{
\small
\begin{equation}
    U_t(m) \;=\; \E_{q,c}\bigl[Y_t \mid \doo{M_t = M_t^{+m}},\, q_t, c_t\bigr]
           \;-\; \E_{q,c}\bigl[Y_t \mid \doo{M_t = M_t^{-m}},\, q_t, c_t\bigr],
    \label{eq:utility}
\end{equation}
}
where $M_t^{+m} = M_t \cup \{m\}$ and $M_t^{-m} = M_t \setminus \{m\}$ denote the memory states (not operations) with and without $m$, respectively, and the expectation marginalizes over the query $q$ and context  $c$.
\vspace{-2mm}
\end{definition}
$U_t(m)$ underlies every memory decision: retaining $m$ if $U_t(m) > 0$.  Section~\ref{sec:identification} shows that $U_t(m)$ factorizes through retrieval into a conditional utility $U_R(m)$, and that the factorization collapses when identification fails. For consolidation, the additive criterion $U_t(m_i)+U_t(m_j)$ ignores interactions between memories. We therefore define consolidation directly through the joint intervention:

{\vspace{-4mm}
\small
\begin{equation}
    \Delta_{\textup{consol}}(m_i,m_j)
    =
    \E\!\left[Y \mid \doo{M_t=(M_t\setminus\{m_i,m_j\})\cup\{\textup{\textsc{Merge}}(m_i,m_j)\}}\right]
    -
    \E\!\left[Y \mid \doo{M_t}\right],
    \label{eq:consol_criterion}
\end{equation}
}
and consolidate when $\Delta_{\textup{consol}}(m_i, m_j) > 0$. Because both terms are expectations over whole memory states, any interaction between $m_i$ and $m_j$ is already inside them, so no additivity is assumed.
We define this criterion so that the operation space is complete, but we do not claim a policy for it. It faces the same identification condition as $U_t(m)$, and Section~\ref{sec:abstention} shows that neither can drive a policy without aggregating over queries in a way we find unavailable.

\vspace{-3mm}
\subsection{Core Challenges}
\label{sec:challenges}
\vspace{-2mm}
The causal formulation reveals four fundamental challenges, annotated in Figure~\ref{fig:causal_graph}:

\begin{enumerate}[label=\textbf{C\arabic*},leftmargin=*,itemsep=0pt]
\vspace{-2mm}

    \item \textbf{Confounded effect estimation.}
The query $q_t$ simultaneously influences the memory operation $O_t$, the retrieved set $R_t$, and the outcome $Y_t$ (red dashed edges in Figure~\ref{fig:causal_graph}), confounding naive comparisons of memory configurations. Without adjustment, performance differences are incorrectly attributed to memory content instead of query difficulty or user intent.

\item \textbf{Support failure through the mediator.}
The intervention $\doo{M_t}$ reaches $Y_t$ only through $R_t$ (thick blue edge in Figure~\ref{fig:causal_graph}). A memory the retriever never surfaces contributes to no output, so $M_t^{+m}$ and $M_t^{-m}$ induce identical retrieved sets, and the two interventional arms become the same experiment. The resulting contrast collapses toward zero, making an unretrieved memory indistinguishable from one with genuinely zero utility. Only the latter justifies $\texttt{forget}$, so a policy cannot distinguish safe from destructive eviction.

\item \textbf{Irreversible operations.}
$\texttt{forget}$ permanently removes a memory, preventing any future re-estimation of its utility. By contrast, $\texttt{noop}$ is reversible: a memory retained in error can be evicted later. Decision rules that treat the two errors symmetrically are therefore misspecified.

\item \textbf{No ground-truth oracle.}
The optimal memory state %$M_t^*$ 
is never observed. Unlike supervised learning, there is no target memory configuration to train against; the value of each memory operation must be inferred solely from its downstream effects.

   \vspace{-2mm}
\end{enumerate}

\vspace{-2mm}
\subsection{Retrieval-Mediated Identification}
\label{sec:identification}
\vspace{-2mm}

Estimating memory utility requires two conditions. \emph{Exchangeability}: timesteps where $m$ is present must be comparable to those where it is absent, so outcome differences are attributable to $m$ rather than to the circumstances under which it appears. \emph{Overlap}: both cases must occur, since an effect cannot be estimated from one alone. Exchangeability is a property of the assignment mechanism, whereas overlap is a property of its support. Retrieval mediation of Section~\ref{sec:scm} determines where overlap must hold, and it is precisely there that it fails. Since $R_t$ mediates the effect of memory on the output, memory utility factorizes into the probability of retrieval and utility conditional on retrieval.

\begin{restatable}[Retrieval-Mediated Factorization]{proposition}{propfactor}
\label{prop:factor}
{Let $\doo{m \in R}$ denote inserting $m$ into the retrieved set $R$, displacing its lowest-ranked memory, and $\doo{m \notin R}$ denote removing $m$ and filling its slot with the highest-ranked memory not already in $R$, so that $|R|$ remains fixed across both arms. Both are dynamic interventions: the displaced or substituted memory depends on the realized ranking at $(M_t,q_t)$.}
Under Definition~\ref{def:scm}, for
any $m$ and any $(q, c)$,

{\vspace{-4mm}
\small
\begin{equation}
\begin{aligned}
    & U(m \mid q, c) = P(m \in R \mid q, c) \cdot U_R(m \mid q, c),
    \quad U(m) = \E_{q,c}\bigl[\, U(m \mid q, c) \,\bigr], \\
    & 
    U_R(m \mid q, c) \triangleq \E\bigl[Y \mid \doo{m \in R},\, q, c\bigr] - \E\bigl[Y \mid \doo{m \notin R},\, q, c\bigr].
    \label{eq:factorization}
    \end{aligned}
\end{equation}
}%
\vspace{-4mm}
\end{restatable}
Proposition~\ref{prop:factor} factorizes memory utility into two terms: \emph{retrievability}, $P(m \in R \mid q,c)$, the probability that $m$ is retrieved for query $q$; and \emph{conditional utility}, $U_R(m \mid q,c)$, the change in expected outcome caused by including $m$ in the retrieved context while holding the context size fixed. Their product is the contribution of $m$ for $(q,c)$, and $U(m)$ averages it over the query distribution.
The factorization also identifies where overlap must hold. Because its first factor is a retrieval probability, overlap is required in the retrieval variable, not in the operation variable.

\begin{assumption}[Retrieval Positivity]
\label{assum:ret_positivity}
For all $m \in M_t$ and all $(q, c)$ in the support:

{\vspace{-4mm}
\small
\begin{equation}
    P\bigl(m \in R_t \mid q_t = q,\, c_t = c\bigr) 
    \;>\; 0.
    \label{eq:ret_positivity}
\end{equation}
}%
\vspace{-6mm}
\end{assumption}
Assumption~\ref{assum:ret_positivity} is the only assumption CMP requires. When it fails, the retrieval probability in Eq.~\ref{eq:factorization} is zero, causing the product to collapse to zero regardless of the conditional utility.
Section~\ref{sec:method} constructs a design in which this assumption holds by construction, with $P(m \in R_t \mid q,c)$ fixed and known. \emph{By contrast, prior methods that estimate memory utility from logged interactions additionally require two conditions (stated as Conditions~\ref{assum:ignorability} and~\ref{assum:positivity} in Appendix~\ref{app:identification}).}

\begin{restatable}[Non-Identifiability under Retrieval Support Failure]{theorem}{propnonid}
\label{prop:nonid}
Fix $(q, c)$ and suppose $P(m \in R \mid q, c) = 0$. Then $f_R(M^{+m}, q) = f_R(M^{-m}, q)$, hence $Y(M^{+m}) = Y(M^{-m})$ almost surely, and $U(m \mid q, c) = 0$ irrespective of $U_R(m \mid q, c)$. Consequently, $U_R(m \mid q,c)$ is not identified by any design that holds $M_{-m}$ fixed while randomizing only whether $m$ is included.
\vspace{-2mm}
\end{restatable}

Theorem~\ref{prop:nonid} separates two failure modes that are observationally indistinguishable: a memory with $U_R(m)=0$ and one with $P(m \in R \mid q,c)=0$ produce the same observed contrast. An estimate near zero therefore admits two interpretations: the memory has no utility, or the design never tests whether it has utility. Only the former justifies $\texttt{forget}$. The same asymmetry applies to $\texttt{store}$: a memory never stored generates no future observations, so its utility can never be estimated.

\vspace{-3mm}
\subsection{Problem Statement}
\vspace{-2mm}
Given logged interactions with a memory-augmented LLM, estimate the conditional utility $U_R(m \mid q,c)$ for every memory a policy may act on. Theorem~\ref{prop:nonid} shows that no design randomizing only the memory state can do so when $P(m \in R \mid q,c)=0$, regardless of sample size or the extent of store-level randomization. \emph{The problem is therefore to construct a design that guarantees Assumption~\ref{assum:ret_positivity} over a specified set of memories and estimates $U_R$ from the resulting data.}

%% file: sections/3_method.tex
\begin{figure}[t]
\centering
\resizebox{\textwidth}{!}{%
\begin{tikzpicture}[
    box/.style={draw, rounded corners=2pt, minimum height=0.6cm, font=\small, thick},
    cell/.style={draw, minimum size=0.3cm, inner sep=0pt},
    det/.style={cell, fill=gray!18},
    exp/.style={cell, fill=blue!30},
    arr/.style={->, >=stealth, thick},
    lbl/.style={font=\scriptsize, align=center},
    ttl/.style={font=\small\bfseries},
]
\node[ttl, anchor=west] at (0, 1.15) {(a) exposure schedule};
\node[ttl, anchor=west] at (6.0, 1.15) {(b) one timestep};
\foreach \r in {0,1,2,3} \foreach \c in {0,...,7} {
    \node[det] at (\c*0.32, -\r*0.32) {};
}
\node[exp] at (0*0.32,0) {};      \node[exp] at (5*0.32,0) {};
\node[exp] at (2*0.32,-0.32) {};  \node[exp] at (6*0.32,-0.32) {};
\node[exp] at (1*0.32,-0.64) {};  \node[exp] at (7*0.32,-0.64) {};
\node[exp] at (3*0.32,-0.96) {};  \node[exp] at (4*0.32,-0.96) {};
\node[lbl, rotate=90] at (-0.42,-0.48) {$m \in \cP$};
\node[lbl, anchor=north] at (1.12,-1.28) {$t = 1, \dots, T$};
\node[lbl, anchor=north, text width=4cm] at (1.12,-1.72)
{each memory receives $n_m$ exposures, $e_m = n_m/T$ is known};

\draw[densely dashed, blue!60] (2.7,-0.48) -- (3.9,-0.48);
\node[lbl, anchor=south, blue!70] at (3.3,-0.38) {supplies $E_t$};

\node[box, minimum width=1.1cm] (store) at (4.6,-0.48) {$M_t$};
\node[box, minimum width=1.8cm, fill=gray!12] (rank) at (6.9,0.15) {top-$(B{-}k)$};
\node[box, minimum width=1.8cm, fill=blue!15] (sched) at (6.9,-1.11) {$E_t$};
\node[box, minimum width=0.9cm] (ctx) at (8.9,-0.48) {$R_t$};
\node[box, minimum width=0.9cm] (gen) at (10.3,-0.48) {$f_Y$};
\node[font=\small] (out) at (11.4,-0.48) {$Y_t$};

\draw[arr] (store) |- (rank);
\draw[arr] (store) |- (sched);
\draw[arr] (rank) -| (ctx);
\draw[arr, blue!70] (sched) -| (ctx);
\draw[arr] (ctx) -- (gen);
\draw[arr] (gen) -- (out);
\node[lbl, anchor=south] at (6.9,0.48) {$B{-}k$ slots, by rank};
\node[lbl, anchor=north] at (6.9,-1.42) {$k$ slots, propensity $e_m$};

\node[ttl, anchor=west] at (13.0, 1.15) {(c) three experimental arms};
\node[lbl, anchor=north, text width=3.0cm] at (14.3,-1.35) {same estimator, different randomization};
\node[box, minimum width=2.3cm] (obs)  at (14.3,0.35) {No randomization};
\node[box, minimum width=2.3cm] (str)  at (14.3,-0.30) {$M_t$ randomized};
\node[box, minimum width=2.3cm, fill=blue!15] (expo) at (14.3,-0.95) {$R_t$ randomized};
\node[lbl, anchor=east] at (13.0,0.35) {observational};
\node[lbl, anchor=east] at (13.0,-0.30) {store-level};
\node[lbl, anchor=east, font=\scriptsize\bfseries] at (13.0,-0.95) {exposure};

\node[box, minimum width=2.3cm] (est) at (17.5,-0.3) {$\hat{U}_R(m),\;\hat{\sigma}(m)$};
\draw[arr, gray!70] (obs.east)  -- (est.west);
\draw[arr, gray!70] (str.east)  -- (est.west);
\draw[arr, blue!70]  (expo.east) -- (est.west);

\end{tikzpicture}
}
\vspace{-6mm}
\caption{Overview of CMP.
(a) Balanced exposure assigns the $Tk$ exposure slots evenly across the exposure pool, fixing each memory's propensity by design.
(b) At each timestep, the context consists of the top-$(B-k)$ retrieved memories together with $k$ randomly exposed memories.
(c) CMP compares three designs that differ only in the randomized variable while using the same estimator.}
\label{fig:cmp_overview}
\vspace{-4mm}
\end{figure}

\vspace{-4mm}
\section{Causal Memory Policy}
\label{sec:method}
\vspace{-2mm}
We develop the \textbf{Causal Memory Policy (CMP)}, which operationalizes retrieval-mediated identification through three components (Figure~\ref{fig:cmp_overview}). Component~(i) introduces a randomized retrieval exposure design in which a fixed number of context slots are withheld from the ranker and filled by memories sampled with known probability. This guarantees Assumption~\ref{assum:ret_positivity} by construction, makes the propensity a design parameter, and removes confounding on the randomized arm. Component~(ii) estimates memory utility by self-normalized inverse-propensity weighting, while including observational and store-level randomized estimators as comparators to isolate the contribution of the design from that of the estimator. Component~(iii) derives the Bayes-optimal decision rule under the asymmetry between reversible and irreversible operations, and shows that every aggregation over queries available to a retention policy is either attenuated or degenerate.

These components address the challenges of Section~\ref{sec:challenges}: Component~(i) resolves \textbf{C1} and \textbf{C2} by randomizing the retrieval mediator; Component~(ii) addresses \textbf{C4} by recovering utility solely from downstream outcomes and separating design effects from estimation effects; and 
Component~(iii) addresses \textbf{C3} by deriving the appropriate decision rule and identifying the remaining obstruction after utility is identified. Theoretical guarantees for each CMP component are established in Appendix~\ref{app:theory}.

\vspace{-4mm}
\subsection{Randomized Retrieval Exposure}
\label{sec:aipw}
\vspace{-2mm}
Theorem~\ref{prop:nonid} shows that randomizing only the memory state cannot identify $U_R(m)$ for memories the retriever never surfaces. The obstruction is that retrieval is deterministic: given $(M_t,q_t)$, the ranker returns a single retrieved set, so a memory outside it is never observed in context and its two potential outcomes are never both realized. Store-level randomization cannot resolve this because retrieval is applied after the store is perturbed. We therefore randomize the mediator itself by reserving a fixed number of retrieval slots for randomized exposure rather than ranking. 
Specifically, At each timestep, the $B$ context slots are filled by two mechanisms:

{\vspace{-4mm}
\small
\begin{equation}
\textbf{Exposure design: }\;
R_t
=
\textup{top-}(B-k)(M_t,q_t)
\;\cup\;
E_t,
\qquad
E_t\subset\mathcal P,\;
|E_t|=k,
\label{eq:exposure}
\end{equation}
}
where $B$ is the context budget and $k\le B$ is the number of slots reserved for exposure. The remaining $B-k$ slots are filled by the deterministic retriever rule of Definition~\ref{def:scm}, while $E_t$ is sampled from an exposure pool $\mathcal P\subseteq M_t$. The extremes recover two familiar designs: $k=0$ gives ordinary retrieval, whereas $k=B$ removes the ranker entirely.

Across $T$ timesteps, the $Tk$ exposure slots are allocated as evenly as possible across $\mathcal P$ and randomly assigned to timesteps. Thus each memory $m\in\mathcal P$ receives exactly
$n_m\in\{\lfloor Tk/|\mathcal P|\rfloor,\lceil Tk/|\mathcal P|\rceil\}$
exposures, and its propensity $e_m=n_m/T$ is known exactly. The propensities are fixed by design rather than estimated, so Assumption~\ref{assum:ret_positivity} holds by construction with $\min_m e_m>0$.

\noindent {\bf Balanced assignment is essential.} Independent inclusion with the same expected count gives $n_m\sim\textup{Binomial}(T,k/|\cP|)$, which leaves many memories underexposed. For example, at $T=90$, $k=2$, and $|\cP|=32$, the expected count is $5.6$, yet roughly one third of the pool receives fewer than five exposures. Balanced assignment instead guarantees every memory exactly five or six exposures. Two properties of the design are worth noting. First, an exposure draw may select a memory already present in the deterministic retrieval, so the realized context contains at most $B$ distinct memories; we report the realized context size in Section~\ref{sec:experiments}. Second, each memory receives $Tk/|\cP|$ exposures, so the horizon required to identify the entire pool grows linearly with $|\cP|/k$ (see Appendix~\ref{app:theory} in detail).

\vspace{-1mm}
\noindent {\bf Exposure pool $\cP$ determines which memories can be identified.} In our experiments, $\cP$ contains the memories relevant to the evaluated queries together with the highest-ranked and a uniform sample of the remaining store, while retrieval continues over the full store. The experiments therefore establish that randomizing the retrieval mediator restores identification, but not that a deployable pool selector captures the memories that matter. Section~\ref{sec:experiments} evaluates three practical selection rules and finds their recall near zero, identifying pool selection as the remaining open problem.

 \vspace{-2mm}
\subsection{Utility Estimation Under Known Propensity}
\label{sec:estimation}
\vspace{-2mm}
We write $Z_m^{(t)}=\ind{m\in R_t}$ for the 
exposure
retrieval indicator with
$\cT_m^{+}=\{t:Z_m^{(t)}=1\}$, 
$\cT_m^{-}=\{t:Z_m^{(t)}=0\}$.
Since $e_m$ is known, conditional utility is estimated by the self-normalized (H\'ajek) inverse-propensity weighting estimator~\citep{imbens2015causal}:

{\vspace{-4mm}
\footnotesize
\begin{equation}
    \hat{U}_{R}(m) \;=\;
    \frac{\sum_{t} Z_m^{(t)} Y_t / e_m}{\sum_{t} Z_m^{(t)} / e_m}
    \;-\;
    \frac{\sum_{t} (1 - Z_m^{(t)}) Y_t / (1 - e_m)}{\sum_{t} (1 - Z_m^{(t)}) / (1 - e_m)}
    \;=\;
    \frac{1}{n_m}\sum_{t \in \cT_m^{+}} Y_t
    -
    \frac{1}{T-n_m}\sum_{t \in \cT_m^{-}} Y_t .
    \label{eq:hajek}
\end{equation}
}
The second equality follows from balanced assignment, which fixes $n_m$ in advance. Consequently, the weight sums are constants, the estimator reduces to a difference of arm means, and its variance,
$\sigma_+^2/n_m+\sigma_-^2/(T-n_m)$, is exact rather than conditional on realized arm sizes. Equation~\ref{eq:hajek} provides both the point estimate and, via Proposition~\ref{prop:hajek}, the standard error $\hat{\sigma}(m)$ used in Section~\ref{sec:abstention}. Self-normalization removes the propensity-dependent scaling of ordinary inverse-propensity weighting, making estimates directly comparable across memories.

\begin{restatable}[H\'ajek Estimator Under Balanced Exposure]{proposition}{prophajek}
\label{prop:hajek}
Under the exposure design of Eq.~\ref{eq:exposure} with balanced assignment,
$\E[\hat{U}_R(m)] = U_R(m)$ for every $m \in \cP$, and
$\textup{Var}(\hat{U}_R(m)) = \sigma_+^2/n_m + \sigma_-^2/(T-n_m)$ exactly,
with $n_m$ fixed by the design.
\vspace{-2mm}
\end{restatable}

We compare CMP with two estimators targeting the same estimand under alternative designs. The observational comparator estimates propensity and outcome models from logged interactions.

{\vspace{-2mm}
\small
\begin{equation}
\begin{aligned}
    & \textbf{Propensity: } \,
    \hat{p}\bigl(m{\in}M \mid M^{-m}, q, c\bigr)
    =
    \text{softmax}\bigl(g_\psi(M^{-m},q,c)\bigr),\\
    & \textbf{Outcome: } \,
    \hat{\mu}\bigl(a,M^{-m},q,c\bigr)
    =
    h_\xi\bigl(\phi(a,M^{-m},q,c)\bigr),
    \label{eq:propensity}
\end{aligned}
\end{equation}
}
where $a \in \{0,1\}$ is the inclusion status, $g_\psi$ is trained by maximum likelihood on logged inclusions, $h_\xi$ is a regression head, and $\psi$ concatenates the inclusion indicator with mean-pooled embeddings of the remaining memories, the query, and the context. Both models condition on $M^{-m}$, which is the adjustment set held fixed in Definition~\ref{def:utility}. 
They are combined using the augmented inverse-propensity weighted (AIPW) estimator~\citep{robins1994estimation, bang2005doubly} 
over the $n$ logged timesteps:

{
\vspace{-6mm}
\footnotesize
\begin{align}
    & \hat{U}_{\textup{obs}}(m) = \frac{1}{n}\sum_{i=1}^{n} \Bigl[
     \bigl(\hat{\mu}(1, M^{-m}_i, q_i, c_i) - \hat{\mu}(0, M^{-m}_i, q_i, c_i)\bigr) \label{eq:aipw_method} \\ & \quad
    + \frac{\ind{m \in M_i}}{\hat{p}(m{\in} \mid M^{-m}_i, q_i, c_i)}
        \bigl(Y_i - \hat{\mu}(1, M^{-m}_i, q_i, c_i)\bigr) 
    - \frac{\ind{m \notin M_i}}{1 - \hat{p}(m{\in} \mid M^{-m}_i, q_i, c_i)}
        \bigl(Y_i - \hat{\mu}(0, M^{-m}_i, q_i, c_i)\bigr)
    \Bigr]   \nonumber
\end{align}
}
Proposition~\ref{prop:dr} establishes consistency for $U(m)$ when either model is correctly specified. 

\begin{restatable}[Double Robustness of the Observational Comparator]{proposition}{propdr}
\label{prop:dr}
%\vspace{-2mm}
$\hat{U}_{\textup{obs}}(m) \xrightarrow{p} U(m)$ if either $\hat{p}$ or $\hat{\mu}$ is correctly specified.
\vspace{-2mm}
\end{restatable}

The \emph{store-level randomized} comparator perturbs store membership, retaining each memory independently with probability $1-p_{\textup{flip}}$, and estimates $\hat{U}_{\textup{exp}}(m)$ by a difference of arm means over timesteps where $m$ is present or absent under the randomization.

\begin{restatable}[Unbiasedness of Store-Level Arm]{proposition}{propshadowunbiased}
\label{prop:shadow_unbiased}
Under independent store-level perturbation with rate $p_{\textup{flip}}$, $\hat{U}_{\textup{exp}}(m)$ is an unbiased 
utility
$
\E_{M^{-m}}
\!\left[
\E[Y\mid m\in M,M^{-m}]
-
\E[Y\mid m\notin M,M^{-m}]
\right],
$
regardless of confounding, with variance
$\sigma_+^2/|\cT_m^+|+\sigma_-^2/|\cT_m^-|$.
\end{restatable}

The three estimators differ only in the variable they randomize: none (observational), memory state $M_t$ (store-level), or retrieved set $R_t$ (exposure). By Theorem~\ref{prop:nonid}, only the exposure design retains support for memories the retriever never surfaces, while all three coincide whenever support is available. This isolates the contribution of the design rather than the estimator, and Section~\ref{sec:experiments} verifies this by restricting the comparison to memories with support under the store-level design. 

\vspace{-2mm}
{\bf Disagreement diagnostic:}
We define
$\Gamma(m)=|\hat{U}_{\textup{obs}}(m)-\hat{U}_{\textup{exp}}(m)|/
(\hat{\sigma}_{\textup{obs}}^2(m)+\hat{\sigma}_{\textup{exp}}^2(m))^{1/2}$,
the standardized disagreement between the observational and store-level arms. Its null is composite: rejection indicates failure of at least one of exchangeability, negligible interaction, or correct model specification. We therefore use $\Gamma$ only as a diagnostic, leaving operational decisions to the uncertainty of the CMP estimate itself (Section~\ref{sec:abstention}).

\vspace{-2mm}
\subsection{From Estimates to Operations}
\label{sec:abstention}
\vspace{-2mm}

An estimate becomes an operation through a decision rule. Challenge \textbf{C3} makes this rule asymmetric: $\texttt{forget}$ permanently removes both the memory and any opportunity to re-estimate its utility, whereas $\texttt{noop}$ is reversible at the next capacity decision. Let $H$ denote the number of future queries over which a retained memory would still contribute, and $\tau$ the number of queries until the decision is revisited, so that $\lambda = H/\tau \geq 1$ is the \emph{irreversibility ratio}. Assuming the posterior $U_R(m)\mid\hat{U}_R,\hat{\sigma}\sim
\mathcal N(\hat{U}_R(m),\hat{\sigma}^2(m))$ with losses $L_{\texttt{forget}}=H\max(0,U_R)$ and $L_{\texttt{noop}}=\tau\max(0,-U_R)$.

\begin{restatable}[One-Sided Abstention is Bayes-Optimal]{theorem}{propabstention}
\label{prop:abstention}
Let $\lambda=H/\tau\ge1$ and $z=\hat{U}_R(m)/\hat{\sigma}(m)$. The Bayes-optimal rule is
$\texttt{forget}(m)\iff g_\lambda(z)<0$,
where $g_\lambda$ is strictly increasing with a unique root
$z^\star(\lambda)$ satisfying
$z^\star(\lambda)<0$ for $\lambda>1$,
$z^\star(1)=0$, and
$z^\star(\lambda)\rightarrow-\infty$
as $\lambda\rightarrow\infty$.
\end{restatable}

{\vspace{-4mm}
\small
\begin{equation}
\texttt{forget}(m)
\iff
\hat{U}_R(m)
<
z^\star(\lambda)\hat{\sigma}(m),
\qquad
g_\lambda(z)
=
\lambda z\Phi(z)
+
z\Phi(-z)
+
(\lambda-1)\phi(z).
\label{eq:onesided}
\end{equation}
}
Here $\phi$ and $\Phi$ denote the standard normal density and distribution functions. Since $g_\lambda$ is strictly increasing, the threshold $z^\star(\lambda)$ is unique. Thus only sufficiently negative estimates trigger eviction. A two-sided rule,
$|\hat{U}_R(m)|>\delta\hat{\sigma}(m)$,
is strictly dominated because it can evict confidently useful memories. When a memory lacks support, $\hat{\sigma}\to\infty$, so $z\to0$ and the rule abstains.

Eq.~\ref{eq:onesided} is nonetheless insufficient to specify a retention policy. The rule requires a scalar $\hat{U}_R(m)$, whereas Proposition~\ref{prop:factor} makes utility query-dependent. Since retention is decided before the next query is known, a policy must aggregate over $\cQ$, but every admissible aggregation is degenerate. Averaging attenuates utility by the fraction of queries requiring the memory (one in thirty on LoCoMo), while $\max_q \hat{U}_R(m\mid q)$ eliminates attenuation but is positive for nearly every memory. Aggregations weighted by similarity to the future query avoid both problems but require information unavailable at decision time. Section~\ref{sec:experiments} evaluates all three and none improves on retaining everything. The same reasoning applies to the other operations: $\texttt{store}$ has the opposite asymmetry, committing unless the estimate is confidently negative, while $\texttt{consolidate}$ inherits the same asymmetry as $\texttt{forget}$ because $\textup{\textsc{Merge}}$ destroys both inputs. Both face the same aggregation problem.

We therefore present CMP as an identified measurement of per-query memory utility rather than a retention policy. On held-out LoCoMo, it separates required from non-required memories for the query on which it is estimated with an AUC of $0.784$, compared with $0.581$ when the same estimator is pooled across queries. The loss in performance shows that the limitation lies not in estimation but in the estimand: a single scalar per memory cannot represent a quantity that varies with the query.

\vspace{-3mm}
\subsection{CMP Algorithm}
\label{sec:algorithm}
\vspace{-2mm}
 
CMP operates in three phases, summarized in Algorithms in Appendix~\ref{app:algorithms}.

(1) {\bf Phase 1: warm-up.}
Collect interaction data under the default policy with the exposure design in Eq.~\ref{eq:exposure}. At each timestep, record $(M_t, q_t, c_t, R_t, Z^{(t)}, Y_t)$, including the realized exposure indicators from which propensities are obtained.

\vspace{-2mm}
(2) {\bf Phase 2: model fitting.}
Estimate $\hat{U}_{R}(m)$ and $\hat{\sigma}(m)$ for each $m\in\cP$ via Eq.~\ref{eq:hajek}, fit propensity and outcome models in Eq.~\ref{eq:propensity}, and compute $\Gamma(m)$ to audit support under the observational comparator.

\vspace{-2mm}
(3) {\bf Phase 3: online execution.}
Continue serving responses while maintaining $\hat{U}_{R}(m\mid q)$ and $\hat{\sigma}(m\mid q)$ online from new interactions. These estimates are consumed by downstream policies: Eq.~\ref{eq:onesided} gives the decision rule for a single query, Section~\ref{sec:abstention} shows why no aggregation over $\cQ$ yields a retention policy, and Section~\ref{sec:experiments} evaluates three candidate aggregations.

%% file: sections/4_experiments.tex
\vspace{-4mm}
\section{Experiments}
\label{sec:experiments}
\vspace{-2mm}

Our experiments answer four questions. \textbf{Q1}: how often does Assumption~\ref{assum:ret_positivity} fail in practice? \textbf{Q2}: can better retrieval repair it, or is randomization necessary? \textbf{Q3}: does randomizing the mediator restore identification, and is the gain due to the design or the estimator? \textbf{Q4}: does identified utility support a retention policy? Section~\ref{sec:main_results} answers Q1, Q2; Section~\ref{sec:exposure_results} answers Q3; and Section~\ref{sec:ablations} answers Q4.

{\bf Setup.} We evaluate on LongMemEval~\citep{wu2024longmemeval}, LoCoMo~\citep{maharana2024evaluating}, HotpotQA~\citep{yang2018hotpotqa}, MuSiQue~\citep{trivedi2022musique}, and Mem0~\citep{chhikara2025mem0}. LongMemEval is our primary benchmark,  LoCoMo supports evaluation under a query distribution, HotpotQA and MuSiQue generalize to multi-hop retrieval, and Mem0 evaluates a deployed memory system. We use \texttt{gemini-3.7-flash} and report token-level F1 and exact match. We do not report LLM-as-judge accuracy because of the reproducibility concerns discussed in Appendix~\ref{app:repro_instruments}. More dataset details are in Appendix \ref{app:datasets}. By default, CMP uses $B=6$ context slots and $|\cP|\approx32$--$36$. Experiment-specific values of $k$ and $T$, together with complete protocols, are given in Appendix~\ref{app:setup}--\ref{app:additional_results}. 

\begin{figure}[t]
\centering

\begin{minipage}[t]{0.47\textwidth}
\footnotesize
\centering
\vspace{-2mm}
\addtolength{\tabcolsep}{-4pt}
\captionof{table}{Retrieval support failure. $P=0$ denotes the fraction of required memories never surfaced by the retriever; \textbf{Store} is the mean number of memories available at query time. $^{\dagger}$Mem0 rewrites source turns, requiring annotated evidence to be matched to stored memories. We report the least-conditioned of three matching rules; see Appendix~\ref{app:protocols_mem0} for details.}
\label{tab:hole}
\begin{tabular}{lccc}
\toprule
\textbf{Setting} & \textbf{Required} & \textbf{$P = 0$} & \textbf{Store} \\
\midrule
LongMemEval (indep.) & $113$ & $\mathbf{54.0\%}$ & $487$ \\
LongMemEval (all)    & $170$ & $61.2\%$          & $487$ \\
LoCoMo               & $359$ & $\mathbf{66.9\%}$ & $588$ \\
Mem0 $2.0.18$$^{\dagger}$ & $109$ & $28.4\%$ & $291$ \\
\bottomrule
\end{tabular}
\end{minipage}
\hfill
\begin{minipage}[t]{0.45\textwidth}
\vspace{-4mm}
\centering
\includegraphics[width=\textwidth]{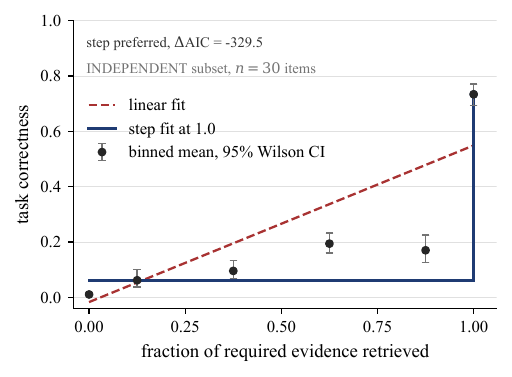}
\vspace{-6mm}
\caption{Task-correctness against the fraction of required evidence retrieved, LongMemEval independent subset.}
\label{fig:step}
\end{minipage}
\vspace{-4mm}
\end{figure}

\vspace{-2mm}
\subsection{Empirical Identification Failure}
\label{sec:main_results}
\vspace{-2mm}
Table~\ref{tab:hole} quantifies identification failure by measuring the fraction of required memories with $P(m\in R\mid q, c)=0$. On LongMemEval, retrieval support is strongly bimodal: $61$ of $113$ are never retrieved and a further $6$ almost never are, $37$ are retrieved on more than $80\%$ of draws, and only $9$ fall between $0.1$ and $0.8$. Consequently, store-level randomization leaves the retrieved set unchanged on $52.2\%$ of interventions on $M_t$, producing $940$ distinct retrieved contexts compared with $3{,}939$ under randomized exposure at the same costs. 

A larger retrieval budget does not resolve the failure. In Mem0, recovering 50\% required memories never surfaced by the retriever requires expanding the retrieval window to $138$ memories, and recovering $90\%$ requires $286$, effectively abandoning retrieval. Mem0 also exposes a separate write-time limitation: across $11{,}651$ memories over $55$ histories, every operation is an addition, with no updates or deletions, so contradictory facts accumulate. This limitation concerns the write policy. See more analysis in Appendix \ref{app:protocols_mem0}. The misses are disproportionately costly. Figure~\ref{fig:step} shows that partial retrieval contributes little: task correctness remains near $0.2$ until every required memory is retrieved, then jumps to $0.73$. The reason is that these tasks require all supporting memories jointly; retrieving only a subset typically leaves the model unable to produce the correct answer.

\vspace{-4mm}
\subsection{Exposure Restores Identification}
\label{sec:exposure_results}
\vspace{-2mm}

\begin{wraptable}{r}{0.45\textwidth}
\vspace{-6mm}
\footnotesize
\centering
\addtolength{\tabcolsep}{-4pt}
\caption{Estimation and decision quality on two designs (only differ in randomized variable) and LongMemEval independent subset.}
\vspace{-2mm}
\label{tab:main}
\begin{tabular}{lccc}
\toprule
\textbf{Design} & \textbf{AUC} & \textbf{Lift} & \textbf{Gold loss} \\
\midrule
Store-level rand.  & $0.542$ & $0.97$ & $10.9$ \\
\textbf{Exposure (CMP)} & $\mathbf{0.664}$ & $\mathbf{0.68}$ & $\mathbf{5.2}$ \\
\bottomrule
\end{tabular}
\vspace{-4mm}
\end{wraptable}
Table~\ref{tab:main} compares store-level randomization and CMP under identical evaluation budgets. We report discrimination (AUC between required and non-required memories), nomination lift (the relative rate at which required memories are nominated for eviction), and gold loss (required memories destroyed per $100$ reclaimed slots under Eq.~\ref{eq:onesided} with $\lambda=10$).

CMP consistently outperforms store-level randomization, improving discrimination from $0.542$ to $0.664$ and halving gold loss from $10.9$ to $5.2$. Moreover, store-level estimates nominate required memories at nearly random (lift $=0.97$), whereas CMP reduces this to $0.68$, making required memories substantially less likely to be selected for eviction. Table~\ref{tab:strata} locates the gap. Stratifying by retrieval support, the two designs are indistinguishable where the store-level design has both arms, and the entire advantage sits in the stratum where it has none. The limitation is therefore the design, not the estimator: store-level estimation is not less accurate.

\begin{wraptable}{r}{0.47\textwidth}
\vspace{-4mm}
\footnotesize
\centering
\addtolength{\tabcolsep}{-4pt}
\caption{Discrimination by retrieval support, LongMemEval independent subset. Median $|t|$ is the store-level estimate's significance.}
\vspace{-2mm}
\label{tab:strata}
\begin{tabular}{lccc}
\toprule
\textbf{Support} & $n$ & \textbf{AUC gap} & \textbf{Median $|t|$} \\
\midrule
$P = 0$            & $61$ & \textbf{exposure only} & $0.83$ \\
$0 < P < 0.8$      & $15$ & $+0.09$ & $1.7$ \\
$P \geq 0.8$       & $37$ & $+0.003$ & $3.66$ \\
\bottomrule
\end{tabular}
\vspace{-4mm}
\end{wraptable}
Appendix~\ref{app:baseline_reported} compares five memory policies on the same axis and finds that heuristic rules nominate required memories at $0.00$--$0.11$ times the chance rate, so the identification failure binds on policies that rank by estimated utility and not on those that do not.
Appendix~\ref{app:examples} provides detailed support-stratified analyses and per-memory case studies (one memory identified only by CMP, one identified by both designs, and one identified by neither).

\vspace{-3mm}
\subsection{Ablation}
\label{sec:ablations}
\vspace{-2mm}

Table~\ref{tab:ablations} separates the design from the decision rule. The two are complementary: holding the decision rule fixed, randomized exposure reduces gold loss from $15$ to $6$, while holding the design fixed, abstention reduces it from $27$ to $6$. Abstention can act on identified memories, whereas randomized exposure restores identification for memories lacking retrieval support. Two-sided abstention performs worse than no abstention, destroying $14.1$ versus $9.0$ required memories per $100$ reclaimed. It admits confidently useful memories into the eviction pool, the behavior ruled out by Theorem~\ref{prop:abstention}.

\begin{wraptable}{r}{0.48\textwidth}
\vspace{-4mm}
\footnotesize
\centering
\addtolength{\tabcolsep}{-5pt}
\caption{Component ablation on LongMemEval. Gold loss is reported per $100$ reclaimed slots.}
\vspace{-2mm}
\label{tab:ablations}
\begin{tabular}{lccc}
\toprule
\textbf{Variant} & \textbf{AUC} & \textbf{Destroyed} & \textbf{Gold loss} \\
\midrule
\textbf{CMP (full)}       & $\mathbf{0.664}$ & $\mathbf{6}$ & $\mathbf{5.2}$ \\
-- one-sided abstention   & $0.664$ & $27$ & $9.0$  \\
-- exposure design        & $0.542$ & $15$ & $10.9$ \\
-- both                   & $0.542$ & $35$ & $11.7$ \\
two-sided abstention      & $0.664$ & {$22$} & {$14.1$} \\
two-sided exposure        & $0.671$ & $10$ & $6.6$  \\
\bottomrule
\end{tabular}
\vspace{-4mm}
\end{wraptable}

Extending the design symmetrically also fails. Adding a forced-exclusion slot closes the remaining support gap but yields only a $0.007$ AUC gain while increasing destroyed memories from $6$ to $10$. Three of the four additional losses are memories the one-sided design could not estimate. Identification is not free: exposing more memories also exposes them to erroneous eviction.

A third variant isolates the effect of the estimand. Averaging utility across queries dilutes the signal because most memories are relevant to a small fraction of queries ($1/30$ for $77$ of $109$ gold memories on LoCoMo). Estimating query-specific utility improves discrimination from $0.581$ to $\mathbf{0.784}$ on the required-memory subset; the pooled estimand over all pool members is $0.551$ (Appendix~\ref{app:results_identification}) and increases the utility gap between required and non-required memories from $0.038$ to $0.300$, confirming that the remaining limitation is in the estimand.

\vspace{-1mm}
\noindent\textbf{Identified utility does not determine retention.}
Per-query utility is accurate for the query on which it is estimated but does not transfer across queries. On a disjoint half of each conversation, no query-independent aggregation achieves correlation above $r=0.10$ with realized contribution, whereas ground-truth %evidence labels reaches 
has $r=0.286$ on the same pairs. This is not a power issue: $99.3\%$ of required $(\text{memory},\text{query})$ cells are identified with five draws per arm. The limitation is structural. A retention policy cannot condition on a future query, while query-independent aggregations either dilute the signal or, in the case of $\max_q \hat{U}_R(m\mid q, c)$, retain essentially every memory. We therefore make no retention claim. CMP provides an identified measurement of per-query memory utility, and our results show that this quantity alone is insufficient to determine what to retain. Appendix~\ref {app:extended_analysis} provides additional analyses of retrieval failure, retrieval scaling, and exposure-budget tradeoffs.

%% file: sections/5_conclusion.tex
\vspace{-3mm}
\section{Conclusion}
\label{sec:conclusion}
\vspace{-3mm}
We presented CMP, a causal framework that treats memory utility as a retrieval-mediated effect. Under deterministic retrieval, many memories are never observed in both treatment arms, leaving their utility unidentifiable from logged interactions. This affects half the required memories on LongMemEval and persists in a deployed memory system. CMP restores identification by randomizing retrieval exposure with known propensity. We prove identification, unbiased estimation, and the optimal decision rule under irreversible operations.

%% file: sections/6_statements.tex
\subsection*{AI use statement}

In this work, we used generative AI tools for writing assistance, including editing and polishing prose, and for code assistance in implementing the experimental harness and analysis scripts. We also used generative AI tools as the generation and embedding backends under study; these are the object of the experiments, and their versions are pinned and reported in Appendix~\ref{app:systems}. We did not use generative AI tools to generate research ideas, to produce experimental results, or to write the theoretical claims or their proofs. All AI-assisted code was reviewed by the authors and its outputs verified against the recorded artifacts; several measurement errors introduced during implementation were found by this process and are reported in Appendix~\ref{app:repro_pool}. We take responsibility for the final content of this work, including all text, claims, and artifacts.

\subsection*{Reproducibility statement}

All theoretical claims are stated with their assumptions in Section~\ref{sec:identification} and proved in Appendix~\ref{app:theory}; Assumption~\ref{assum:ret_positivity} is the only condition our method requires, and the conditions needed by the comparator estimators are stated separately in Appendix~\ref{app:identification}. Benchmark construction, the item selection and its exclusions, model versions, and the exposure design parameters for every experiment are given in Appendix~\ref{app:setup}, and the measurement protocol for each reported quantity in Appendix~\ref{app:protocols}. Appendix~\ref{app:algorithms} specifies the exposure allocation and estimation procedures. Appendix~\ref{app:reproducibility} records the discrepancies we found in our own apparatus, including a metadata leak in the repository's default serving path that does not affect the reported arms, and the instability of the LLM-judge protocol, which is why we report only deterministic metrics. Assignment permutations, item selections, and pool constructions are seeded and the seeds are stored with the results they produced. Code, per-item outputs, and scores are available at \url{https://anonymous.4open.science/r/cmp-release-D0C3/}.

\subsection*{Ethics statement}

This work uses only publicly available benchmarks and involves no human subjects, no personally identifying information, and no data collection. The conversational benchmarks we evaluate contain synthetic or released dialogue, and we introduce no new data. Our method randomizes which stored memories are placed in a model's context, which changes the information available to the system on a fraction of interactions; in a deployed setting this could degrade individual responses, and Section~\ref{sec:experiments} reports the measured cost. Deployments should weigh that cost against the value of the estimates the design produces. We report a failure mode in a third-party memory system, and we do so to characterize a condition that applies to the class of systems.

%% file: appendix/1_related_work.tex
\section{Related Work}
\label{app:related}

\vspace{-2mm}
\noindent{\bf Memory-augmented LLMs.}
Retrieval-augmented generation~\citep{lewis2020retrieval} establishes the paradigm of conditioning LLM outputs on passages retrieved from an external store. MemoryBank~\citep{zhong2024memorybank} introduces Ebbinghaus-inspired forgetting curves for long-term retention; generative agents~\citep{park2023generative} maintain hierarchical memory with reflection in simulated social environments; and memorizing transformers~\citep{wu2022memorizing} embed persistent key--value memory directly into attention layers. More recent work has augmented LLMs with unbounded long-term stores~\citep{wang2023augmenting}, evaluated memory over extended conversations~\citep{maharana2024evaluating}, and developed agentic memory management. In particular, A-Mem~\citep{xu2025mem} draws on the Zettelkasten method to autonomously organize memories through note construction and link generation. Despite these advances, memory operations are still largely governed by hand-designed mechanisms, e.g., similarity thresholds, recency scores, and capacity constraints, rather than a formal framework for determining when and why a memory operation improves downstream performance.

\noindent{\bf Estimating per-memory utility.}
Recent systems estimate the contribution of each memory rather than assuming it. Memory governance based on per-memory success and failure counts~\citep{simsek2026forget} converges to a conditional success probability; paired trajectory auditing~\citep{gao2026skillaudit} isolates a candidate's effect by executing with and without it; budget-curated retention~\citep{wu2026forget} selects memories under an explicit capacity constraint; and causal memory selection~\citep{srivastava2026causal} scores candidates through controlled interventions on the candidate set. These approaches differ in their estimators and supervision, but share a common dependency: evidence about a memory can be collected only when the retriever surfaces it. Our contribution is  orthogonal to the particular utility estimator: the retrieval-support condition we identify is a prerequisite for all of them. Moreover, causal memory selection~\citep{srivastava2026causal} requires access to the ground-truth answer within the selection loop, an additional source of supervision that is generally unavailable at deployment.

\noindent{\bf Memory policy optimization.}
MemGPT~\citep{packer2023memgpt} draws on operating-system virtual memory to let LLMs page information between context and external storage, but relies on the LLM's own judgment for management decisions without formal optimization. More recently, MemPO~\citep{li2026mempo} proposes self-memory policy optimization, framing memory as a learnable action within an RL loop and introducing trajectory-level and memory-level advantage estimation for credit assignment. MemPO achieves strong empirical gains on long-horizon benchmarks by training the agent to autonomously compress and retain information. However, RL-based approaches optimize memory decisions end-to-end without asking whether the effect of an individual memory is identified from the data the system collects. The reward signal may be observed even when the per-memory contrast needed to attribute that reward is not. Our work identifies a prior condition: the effect of a specific memory can be identified from retrieval-conditioned observations only where the retriever provides support for that memory. Where retrieval provides no such support, end-to-end optimization cannot by itself recover the missing per-memory contrast.

\noindent{\bf Memory consolidation in cognitive science.}
Complementary learning systems theory~\citep{kumaran2016learning, mcclelland2020integration} posits that biological memory relies on two interacting systems: a fast hippocampal store for episodic details and a slow neocortical store for consolidated abstractions. Sleep-dependent consolidation selectively strengthens, integrates, and prunes memories according to their future relevance~\citep{stickgold2005sleep}. Prior AI work~\citep{torre2025bio, dehghani2024bio, ezeji2025towards} has drawn bio-inspired heuristics from these theories, but does not formally characterize when consolidation improves downstream performance. We instead formulate consolidation as a joint intervention (Eq.~\ref{eq:consol_criterion}), making explicit the causal quantity that a consolidation decision would require. This quantity remains subject to the same identification condition as the single-memory estimand; accordingly, our criterion characterizes the effect of consolidation rather than prescribing a consolidation policy.

\noindent{\bf Causal inference in machine learning.}
Structural causal models (SCMs)~\citep{pearl2009causality} provide the formal language for reasoning about interventions ($do$-calculus) and counterfactuals. This causal perspective has yielded principled methods across several areas: counterfactual fairness~\citep{kusner2017counterfactual} uses SCMs to define equitable predictions; causal representation learning~\citep{scholkopf2021toward} seeks disentangled latent factors respecting the data-generating process, including when the causal structure among them is not assumed~\citep{behnam2026reasoning}; and inverse propensity weighting~\citep{rosenbaum1983central} underpins treatment-effect estimation and off-policy evaluation. Doubly robust estimators~\citep{robins1994estimation, wang2024handling} combine outcome modeling with propensity weighting, providing consistency when either model is correctly specified while reducing variance of pure IPW. Recent work has begun integrating causal reasoning into LLM-based agents: Causal-aware LLMs~\citep{hu2025causal} embed SCMs into RL agents through a learning--adapting--acting paradigm for environment modeling, while REMI~\citep{raman2025remi} constructs personal causal knowledge graphs for lifestyle recommendation. Neither, however, applies causal inference to memory management itself. Doing so requires more than transplanting standard causal machinery: as Section~\ref{sec:identification} shows, identification depends critically on the retrieval process, whose support determines which memory effects can be learned from the observed data.

%% file: appendix/2_theory.tex
\section{Notation}
\label{app:notation}

Table~\ref{tab:notation} summarizes the notation used throughout the paper.

\begin{table}[h]
\centering
\small
\caption{Summary of notation used throughout the paper.}
\label{tab:notation}
\begin{tabular}{ll}
\toprule
\textbf{Symbol} & \textbf{Description} \\
\midrule
$\cM = (\cU, \cV, \cF)$ & Structural causal model (Def.~\ref{def:scm}) \\
$M_t$ & Memory state at time $t$ \\
$M^{-m}_t$ & Memory state excluding $m$, $M_t \setminus \{m\}$ \\
$\mathbb{M}$ & Universe of possible memories \\
$R_t$ & Retrieved subset at time $t$ \\
$O_t$ & Memory operation at time $t$ \\
$\cO$ & Operation space $\{\texttt{store}, \texttt{forget}, \texttt{consolidate}, \texttt{noop}\}$ \\
$Y_t$ & Task output at time $t$ \\
$q_t, c_t$ & Query and observed context at time $t$ \\
$\cQ$ & Task distribution over queries \\
$\theta_{\textup{LLM}}$ & Frozen LLM parameters \\
\midrule
$U_t(m)$ & Memory utility at time $t$ (Def.~\ref{def:utility}); written $U(m)$ when unambiguous \\
$U_R(m \mid q, c)$ & Conditional utility given retrieval (Eq.~\ref{eq:factorization}) \\
$\Delta_{\textup{consol}}(m_i, m_j)$ & Consolidation criterion (Eq.~\ref{eq:consol_criterion}) \\
\midrule
$B$ & Context budget, number of memories the generator receives \\
$k$ & Slots withheld from the ranker per timestep, $k \leq B$ \\
$\cP$ & Exposure pool, memories whose utility is estimated \\
$T$ & Horizon, number of estimation draws \\
$E_t$ & Exposed set at time $t$ (Eq.~\ref{eq:exposure}) \\
$Z^{(t)}_m$ & Retrieval indicator $\ind{m \in R_t}$, randomized by the design \\
$n_m$ & Exposures allocated to $m$ under balanced assignment \\
$e_m$ & Known exposure propensity, $e_m = n_m / T$ \\%(Eq.~\ref{eq:propensity_known}) \\
$\cT^{+}_m, \cT^{-}_m$ & Timesteps with $m$ exposed and not exposed \\
\midrule
$\hat{U}_R(m)$ & H\'ajek estimate of $U_R(m)$ (Eq.~\ref{eq:hajek}) \\
$\hat{\sigma}(m)$ & Standard error of $\hat{U}_R(m)$ (Prop.~\ref{prop:hajek}) \\
$\sigma_+^2, \sigma_-^2$ & Outcome variance within the exposed and unexposed arms \\
$\hat{U}_{\textup{obs}}(m)$ & Observational (AIPW) comparator (Eq.~\ref{eq:aipw_method}) \\
$\hat{U}_{\textup{exp}}(m)$ & Store-level randomized comparator \\
%$U_{\textup{marg}}(m)$ & Marginal utility targeted by the store-level arm (Prop.~\ref{prop:shadow_unbiased}) \\
$p_{\textup{flip}}$ & Store-level perturbation rate \\
$\hat{p}, \hat{\mu}$ & Propensity and outcome models (Eq.~\ref{eq:propensity}) \\
$\Gamma(m)$ & Disagreement diagnostic, reported as a support audit \\
$\psi$ & Feature map for the outcome model (Eq.~\ref{eq:propensity}) \\
\midrule
$H, \tau$ & Horizon and interval to the next decision \\
% $\lambda$ & Irreversibility ratio $H / \tau$ \\
$g_\lambda$ & Threshold function of Eq.~\ref{eq:onesided} \\
$z^\star(\lambda)$ & Abstention threshold, the root of $g_\lambda$ \\
$\phi, \Phi$ & Standard normal density and distribution function \\
$\epsilon_p$ & Propensity clipping threshold for the observational comparator, $0.01$ \\
$n_{\min}$ & Minimum draws per arm for a memory to count as identified \\
$\Lambda$ & Sensitivity parameter for unmeasured confounding \\
\bottomrule
\end{tabular}
\end{table}

\section{Theoretical Guarantees}
\label{app:theory}
This appendix collects the formal statements and proofs supporting Sections~\ref{sec:identification} and~\ref{sec:method}. Appendix~\ref{app:factorization_proofs} proves the factorization and non-identifiability results that motivate the exposure design; Appendix~\ref{app:hajek_proof} gives the estimator's properties and the budget it requires; Appendix~\ref{app:abstention_proof} proves the decision rule under irreversible operations; and Appendix~\ref{app:identification} states the conditions required by the observational comparator, which CMP does not need.

\subsection{Retrieval-Mediated Identification}
\label{app:factorization_proofs}
 
This subsection proves the two results of Section~\ref{sec:identification}. Both follow from the single structural fact that $Y_t$ depends on $M_t$ only through $R_t$, and neither requires an assumption beyond Definition~\ref{def:scm} and the slot-substitution convention of Proposition~\ref{prop:factor}, which fixes what $\doo{m \notin R}$ means.
 
\propfactor*
 
\begin{proof}
\label{app:proof_factor}
Fix $(M^{-m}, q, c)$ and suppress $M^{-m}$ in the notation throughout, since it is held fixed. Write $A = \ind{m \in R}$ for the retrieval indicator under the observational regime, so $\Pr(A = 1) = P(m \in R \mid q, c)$. By
Definition~\ref{def:scm}, $Y = f_Y(R, q, \theta_{\textup{LLM}})$, so $Y$ is a
deterministic function of $R$ given $(q, \theta_{\textup{LLM}})$ and the
intervention $\doo{M = M^{+m}}$ acts on $Y$ only by changing $R$.
 
Consider the two states $M^{+m}$ and $M^{-m}$ of Definition~\ref{def:utility}.
Under the slot-substitution convention stated in
Proposition~\ref{prop:factor}, $|R|$ is held fixed across the two arms, so the
retrieved sets differ in at most the single element $m$: either $m \in
f_R(M^{+m}, q)$ and $f_R(M^{-m}, q)$ replaces it by the highest-ranked memory
not already retrieved, or $m \notin f_R(M^{+m}, q)$ and the two retrieved sets
coincide. Partitioning on which case obtains,
\begin{align*}
    &\E\bigl[Y \mid \doo{M = M^{+m}}, q, c\bigr] - \E\bigl[Y \mid \doo{M = M^{-m}}, q, c\bigr] \\
    & = \Pr(A = 1)\cdot\Bigl(\E\bigl[Y \mid \doo{m \in R}, q, c\bigr] - \E\bigl[Y \mid \doo{m \notin R}, q, c\bigr]\Bigr) \\
    &\quad + \Pr(A = 0)\cdot 0,
\end{align*}
since on the event $A = 0$ the two interventions induce the same retrieved set
and therefore the same outcome. The second factor in the first term is
$U_R(m \mid q, c)$ by definition, giving
\begin{equation*}
    U(m \mid q, c) = P(m \in R \mid q, c)\cdot U_R(m \mid q, c).
\end{equation*}
Taking the expectation over $(M^{-m}, q, c)$ and applying
Definition~\ref{def:utility} yields Eq.~\ref{eq:factorization}.
\end{proof}

\begin{remark}[Store-level randomization does not repair this at feasible horizons]
\label{rem:exponential}
A design that perturbs the entire store rather than $m$ alone can in principle surface $m$: under independent retention at rate $1 - p_{\textup{flip}}$, a memory at rank $r$ enters the top-$B$ whenever at least $r - B$ of the $r-1$ higher-ranked memories are dropped, an event of probability $\Pr[\textup{Bin}(r-1, p_{\textup{flip}}) \geq r - B] > 0$. The obstruction is therefore overlap rather than identification in the strict sense. It is nonetheless binding: at $p_{\textup{flip}} = 0.1$, $B = 6$, and $r = 23$, this probability is below $10^{-8}$, so observing a single inclusion requires a horizon beyond any feasible budget. Section~\ref{sec:experiments} measures the consequence: $52.2\%$ of store perturbations left the retrieved set unchanged, and no perturbation produced an inclusion for the memories of Table~\ref{tab:hole}.
\end{remark}

\propnonid*
 
\begin{proof}
\label{app:proof_nonid}
Fix $(M^{-m}, q, c)$ with $P(m \in R \mid q, c) = 0$. Then $m \notin f_R(M^{+m}, q)$
almost surely, and by the slot-substitution convention $f_R(M^{-m}, q)$ differs
from $f_R(M^{+m}, q)$ only by removing $m$ and backfilling; since $m$ was not
present, no removal occurs and
\begin{equation*}
    f_R(M^{+m}, q) = f_R(M^{-m}, q) \quad\text{almost surely.}
\end{equation*}
Because $Y = f_Y(R, q, \theta_{\textup{LLM}})$ is a deterministic function of
$R$ given $(q, \theta_{\textup{LLM}})$, the two arms produce identical outcomes,
so $U(m \mid q, c) = 0$. This also follows from
Proposition~\ref{prop:factor} with the leading factor equal to zero, and holds
for every value of $U_R(m \mid q, c)$.
 
For the second claim, let $\mathcal{D}$ be any design that assigns membership of $M_t$ by a known or unknown randomization, and let $\Pi$ denote the induced distribution over observed trajectories $(M_t, R_t, q_t, c_t, Y_t)$. Every realization of $\mathcal{D}$ determines $R_t = f_R(M_t, q_t)$ through the same deterministic map, so for any two memory states differing only in $m$ the observed $R_t$ and hence $Y_t$ are identical. 

The restriction to $(q, c)$ matters here. A design may well surface $m$ at some other query $(q', c')$ and thereby identify $U_R(m \mid q', c')$, but $U_R$ is a conditional quantity: observations at $(q', c')$ place no constraint on its value at $(q, c)$ without a further assumption relating the two, which Definition~\ref{def:scm} does not supply. Consequently $\Pi$ is invariant to $U_R(m \mid q, c)$: two data-generating processes agreeing in every respect except the value of $U_R(m \mid q, c)$ induce the same distribution over observables. No functional of $\Pi$ can therefore distinguish them, which is the definition of non-identifiability. The argument does not use the form of the randomization and applies in particular when membership of $M_t$ is randomized with propensity bounded away from zero and one.
\end{proof}
 
\begin{remark}[Why the failure is invisible]
\label{rem:invisible}
Theorem~\ref{prop:nonid} concerns identification, not estimation. An estimator applied to such a memory does not fail to return a value; it returns a difference of two sample means drawn from the same distribution, which concentrates at zero. The estimate is therefore indistinguishable from that of a memory with $U_R(m) = 0$, and the two are separated only by the design's support, which the estimate itself does not report.
\end{remark}

In the strongest case the two arms contain byte-identical contexts on every draw, and the estimate is exactly zero with zero variance; Appendix~\ref{app:examples} gives such a memory.
 
\subsection{Estimation Under the Exposure Design}
\label{app:hajek_proof}
 
\prophajek*
 
\begin{proof}
\label{app:proof_hajek}
Write $Z^{(t)}_m = \ind{m \in R_t}$. Balanced round-robin assignment (Algorithm~\ref{alg:assign}) fixes the number of exposures $n_m = \sum_{t=1}^{T} Z^{(t)}_m$ in advance, with $e_m = n_m / T$ known. Two consequences follow immediately and are used throughout.
 
\emph{The arm sizes are deterministic.} $|\cT^{+}_m| = n_m$ and $|\cT^{-}_m| =
T - n_m$ hold with probability one, not in expectation. The randomness in the
design lies entirely in \emph{which} timesteps receive the exposure, drawn by a
uniformly random permutation of the assignment, and not in how many.
 
\emph{The self-normalized weights collapse.} Since $Z^{(t)}_m \in \{0,1\}$ and
$e_m$ does not vary with $t$,
\begin{equation*}
    \frac{\sum_t Z^{(t)}_m Y_t / e_m}{\sum_t Z^{(t)}_m / e_m}
    = \frac{\sum_t Z^{(t)}_m Y_t}{\sum_t Z^{(t)}_m}
    = \frac{1}{n_m}\sum_{t \in \cT^{+}_m} Y_t ,
\end{equation*}
and symmetrically for the excluded arm with $T - n_m$ in place of $n_m$. This
is the second equality of Eq.~\ref{eq:hajek}.
 
\textbf{Unbiasedness.} By construction the permutation is drawn independently of
$(q_t, c_t)$ and of the outcome mechanism, so $Z^{(t)}_m \perp\!\!\!\perp (q_t,
c_t)$. Writing $\mu_1 = \E[Y_t \mid Z^{(t)}_m = 1]$ and $\mu_0 = \E[Y_t \mid
Z^{(t)}_m = 0]$, each term of the first arm satisfies $\E[Y_t \mid t \in
\cT^{+}_m] = \mu_1$, and the arm is an average of $n_m$ such terms, so
$\E\bigl[\tfrac{1}{n_m}\sum_{t \in \cT^{+}_m} Y_t\bigr] = \mu_1$. The same
holds for the second arm with $\mu_0$. Hence
\begin{equation*}
    \E\bigl[\hat{U}_R(m)\bigr] = \mu_1 - \mu_0
    = \E\bigl[Y \mid \doo{m \in R}\bigr] - \E\bigl[Y \mid \doo{m \notin R}\bigr]
    = U_R(m),
\end{equation*}
where the second equality holds because the exposure indicator is assigned by
the design and is therefore independent of all other variables in $\cM$, so
conditioning on it coincides with intervening on it. No assumption on
confounding between $(q_t, c_t)$ and the memory state is used.
 
\textbf{Variance.} In the estimation regime, the memory state is held fixed across draws, so outcomes at distinct draws are independent given the design; under a sequentially evolving store, the potential outcomes are auto-correlated through $M_t$, and the expression below is a lower bound. Writing $\sigma_+^2 =
\textup{Var}(Y_t \mid Z^{(t)}_m = 1)$ and $\sigma_-^2 = \textup{Var}(Y_t \mid
Z^{(t)}_m = 0)$,
\begin{equation*}
    \textup{Var}\bigl(\hat{U}_R(m)\bigr)
    = \textup{Var}\Bigl(\tfrac{1}{n_m}\sum_{t \in \cT^{+}_m} Y_t\Bigr)
    + \textup{Var}\Bigl(\tfrac{1}{T - n_m}\sum_{t \in \cT^{-}_m} Y_t\Bigr)
    = \frac{\sigma_+^2}{n_m} + \frac{\sigma_-^2}{T - n_m}.
\end{equation*}
The denominators are constants fixed by the design, so this expression is exact
and not conditional on realized arm sizes. Under independent inclusion with
$\Pr(Z^{(t)}_m = 1) = e_m$ the arm sizes are $\textup{Binomial}(T, e_m)$ and
the corresponding statement holds only conditionally on them, with an
additional term from their variability; balanced assignment removes it.
 
The standard error used in Section~\ref{sec:abstention} is
$\hat{\sigma}(m) = \bigl(\hat{\sigma}_+^2/n_m + \hat{\sigma}_-^2/(T -
n_m)\bigr)^{1/2}$ with $\hat{\sigma}_\pm^2$ the within-arm sample variances,
and is defined whenever $n_m \geq 2$ and $T - n_m \geq 2$.
\end{proof}
 
\begin{corollary}[Exposure Budget]
\label{cor:budget}
Under balanced assignment with budget $B$, exposed slots $k$, and pool $\cP$, every $m \in \cP$ receives $n_m \in \{\lfloor Tk/|\cP| \rfloor, \lceil Tk/|\cP| \rceil\}$ exposures. Attaining a minimum of $n_{\min}$ draws in each arm for every pool member therefore requires
\begin{equation}
    T \;\geq\; n_{\min} \cdot \frac{|\cP|}{k},
    \label{eq:budget}
\end{equation}
so the horizon grows linearly in $|\cP|/k$ and does not depend on $|M_t|$.
\end{corollary}
 
\begin{proof}
\label{app:proof_budget}
By construction the $Tk$ exposure slots are divided as evenly as possible among $|\cP|$ members, so $n_m \geq \lfloor Tk/|\cP| \rfloor$ for every $m$. Requiring $\lfloor Tk/|\cP| \rfloor \geq n_{\min}$ gives
Eq.~\ref{eq:budget}. The excluded arm satisfies $T - n_m \geq n_{\min}$ when $\lceil Tk/|\cP| \rceil \leq T - n_{\min}$, that is when $k/|\cP| < 1$ with margin. When $k$ is large relative to $|\cP|$ a memory may be exposed on nearly every timestep and the excluded arm starves; Section~\ref{sec:ablations} reports this case.
\end{proof}

\begin{remark}[What the budget does not buy]
\label{rem:budget_scope}
Eq.~\ref{eq:budget} bounds the horizon required to identify the members of $\cP$, not the full store $M_t$. Identifying additional memories requires enlarging $\cP$, increasing the required horizon proportionally. The budget therefore trades coverage for horizon, and the choice of $\cP$ determines which memories can be identified.
\end{remark}

\subsection{Decision Rules Under Irreversibility}
\label{app:abstention_proof}

Section~\ref{sec:abstention} asks what an identified estimate supports as an operation. The answer turns on an asymmetry in the losses rather than on the estimate's precision: $\texttt{forget}$ destroys the memory and the channel through which its utility could be re-estimated, while $\texttt{noop}$ is revisited at the next decision point. The following result characterizes the Bayes rule under that asymmetry.

\propabstention*
 
\begin{proof}
\label{app:proof_abstention}
Under the posterior $U \sim \cN(\hat{U}_R(m), \hat{\sigma}^2(m))$, write
$z = \hat{U}_R(m)/\hat{\sigma}(m)$ and use the standard truncated-normal
moments
\begin{equation*}
    \E[\max(0, U)] = \hat{\sigma}\bigl[z\Phi(z) + \phi(z)\bigr],
    \qquad
    \E[\max(0, -U)] = \hat{\sigma}\bigl[-z\Phi(-z) + \phi(z)\bigr].
\end{equation*}
Eviction is preferred iff $\E[L_{\texttt{forget}}] < \E[L_{\texttt{noop}}]$,
that is iff
\begin{equation*}
    H\hat{\sigma}\bigl[z\Phi(z) + \phi(z)\bigr] < \tau\hat{\sigma}\bigl[-z\Phi(-z) + \phi(z)\bigr].
\end{equation*}
Dividing by $\tau\hat{\sigma} > 0$ and setting $\lambda = H/\tau$, this is
$\lambda[z\Phi(z) + \phi(z)] + z\Phi(-z) - \phi(z) < 0$, which is
$g_\lambda(z) < 0$ for $g_\lambda$ as in Eq.~\ref{eq:onesided}.
 
\textbf{(i) Monotonicity.} Using $\phi'(z) = -z\phi(z)$,
\begin{align*}
    g_\lambda'(z)
    &= \lambda\Phi(z) + \lambda z\phi(z) + \Phi(-z) - z\phi(z) - (\lambda-1)z\phi(z) \\
    &= \lambda\Phi(z) + \Phi(-z) + z\phi(z)\bigl[\lambda - 1 - \lambda + 1\bigr]
     = \lambda\Phi(z) + \Phi(-z) \;>\; 0 .
\end{align*}
 
\textbf{(ii) Existence, uniqueness, and sign of the root.} $g_\lambda(0) =
(\lambda - 1)\phi(0) > 0$ for $\lambda > 1$, and $g_\lambda(z) \to -\infty$ as
$z \to -\infty$ since $z\Phi(-z) \to -\infty$ while the remaining terms are
bounded. Continuity and (i) give a unique root, and $g_\lambda(0) > 0$ places
it strictly below zero. The rule $g_\lambda(z) < 0$ is therefore
$z < z^\star(\lambda)$, which is Eq.~\ref{eq:onesided}.
 
\textbf{(iii) The symmetric case.} At $\lambda = 1$, $g_1(z) = z[\Phi(z) +
\Phi(-z)] = z$, whose unique root is $z = 0$; the rule reduces to thresholding
on the sign of $\hat{U}_R(m)$ and abstention disappears.
 
\textbf{(iv) Monotonicity in $\lambda$.} $\partial g_\lambda/\partial\lambda =
z\Phi(z) + \phi(z) = \E[\max(0,U)]/\hat{\sigma} > 0$ for every $z$, so
$g_\lambda$ increases pointwise in $\lambda$ and its root moves left; strict
monotonicity of $z^\star$ follows from (i) and the implicit function theorem.
Since $z\Phi(z) + \phi(z) > 0$ everywhere, for any fixed $z$ there is a
$\lambda$ beyond which $g_\lambda(z) > 0$, so no finite $z$ remains in the
eviction region as $\lambda \to \infty$.
\end{proof}
 
\begin{corollary}[Two-sided rules are dominated]
\label{cor:twosided}
For $\lambda > 1$, any rule that evicts some $m$ with $\hat{U}_R(m) > 0$ incurs
strictly greater expected loss than abstaining on that $m$.
\end{corollary}
 
\begin{proof}
\label{app:proof_twosided}
If $\hat{U}_R(m) > 0$ then $z > 0$, and by (i)--(ii) of Theorem~\ref{prop:abstention}, $g_\lambda(z) > g_\lambda(z^\star) = 0$.
Since $g_\lambda(z) < 0$ characterizes the region where eviction has lower
expected loss, eviction is strictly worse than $\texttt{noop}$ at every such
$z$. A rule of the form $|\hat{U}_R(m)| > \delta\hat{\sigma}(m)$ evicts on
$z > \delta$ and is therefore dominated for every $\delta > 0$.
\end{proof}
 
\begin{remark}[Behaviour under failed identification]
\label{rem:abstain_unidentified}
When a memory has no support, $\hat{\sigma}(m)$ is unbounded and $z \to 0$, so
$g_\lambda(z) \to (\lambda - 1)\phi(0) > 0$ and the rule abstains. The rule
therefore degrades to $\texttt{noop}$ exactly on the memories of
Theorem~\ref{prop:nonid}, without requiring a separate diagnostic to detect
them.
\end{remark}
 
\begin{remark}[Calibration]
\label{rem:calibration}
Solving $g_\lambda(z^\star) = 0$ numerically gives $z^\star(10) \approx -0.90$,
$z^\star(100) \approx -1.72$, and $z^\star(1000) \approx -2.42$. The threshold
grows logarithmically in $\lambda$, so even a thousand-fold irreversibility
penalty calls for roughly $2.4$ standard errors.
\end{remark}

\subsection{Conditions Required by the Observational Comparator}
\label{app:identification}

The observational comparator of Section~\ref{sec:estimation} estimates $U(m)$ from logged interactions without randomization, and therefore requires two conditions the exposure design does not.

\begin{comparatorcond}[Inclusion-Level Ignorability]
\label{assum:ignorability}
Conditional on $(M^{-m}_t, q_t, c_t)$, the inclusion indicator $\ind{m \in M_t}$ is independent of the potential outcomes: $Y_t(a) \perp\!\!\!\perp \ind{m \in M_t} \mid M^{-m}_t, q_t, c_t$ for $a \in \{0,1\}$.
\end{comparatorcond}

The assumption is stated at the level of the treatment the estimator uses. Inclusion is a deterministic function of the operation history, so per-step ignorability for $O_t$ does not by itself deliver exchangeability for $\ind{m \in M_t}$.

\begin{comparatorcond}[Feasibility-Restricted Positivity]
\label{assum:positivity}
For all feasible operations $o \in \cO(M_t)$ and all $(M, q, c)$ in the support, $p(O_t = o \mid M_t = M, q_t = q, c_t = c) > 0$.
\end{comparatorcond}

Restricting to the feasible set $\cO(M_t)$ avoids requiring nonzero probability for structurally impossible operations. This is the overlap condition prior work verifies, and it is independent of Assumption~\ref{assum:ret_positivity}.

Under these two conditions the observational estimator is consistent for $U(m)$ if either nuisance model is correctly specified.
\propdr*

\begin{proof}
\label{app:proof_dr}
We prove double robustness by showing $\E[\hat{U}_{\textup{obs}}(m)] = U(m)$ when either $\hat{p} = p$ or $\hat{\mu} = \mu$.

Define the treatment indicator $A_i = \ind{m \in M_i}$ and the shorthand $e_i = p(m \in M_i \mid M^{-m}_i, q_i, c_i)$ for the true propensity, $\hat{e}_i = \hat{p}(m \in M_i \mid M^{-m}_i, q_i, c_i)$ for the estimated propensity, $\mu_1(M^{-m}, q, c) = \E[Y \mid m \in M, M^{-m}, q, c]$, and $\mu_0(M^{-m}, q, c) = \E[Y \mid m \notin M, M^{-m}, q, c]$. The AIPW estimator (Eq.~\ref{eq:aipw_method}) can be written as:
\begin{equation*}
\begin{aligned}
 \hat{U}_{\textup{obs}}(m) & = \frac{1}{n}\sum_{i=1}^n \left[
        \bigl(\hat{\mu}_1(M^{-m}_i, q_i, c_i) - \hat{\mu}_0(M^{-m}_i, q_i, c_i)\bigr)\right] \\
  & \quad     + \frac{1}{n}\sum_{i=1}^n \left[ \frac{A_i}{\hat{e}_i}(Y_i - \hat{\mu}_1(M^{-m}_i, q_i, c_i))
        - \frac{1 - A_i}{1 - \hat{e}_i}(Y_i - \hat{\mu}_0(M^{-m}_i, q_i, c_i))
    \right].
\end{aligned}
\end{equation*}

\textbf{Case 1: $\hat{\mu} = \mu$ (outcome model correct).}
When $\hat{\mu}_1 = \mu_1$ and $\hat{\mu}_0 = \mu_0$, the residuals $Y_i - \hat{\mu}_1(M^{-m}_i, q_i, c_i)$ have conditional mean zero given $A_i = 1, q_i, c_i$. Thus:
\begin{equation*}
    \E\!\left[\frac{A_i}{\hat{e}_i}(Y_i - \mu_1(M^{-m}_i, q_i, c_i)) \;\Big|\; q_i, c_i\right] 
    = \frac{e_i}{\hat{e}_i} \cdot \E[Y_i - \mu_1(M^{-m}_i, q_i, c_i) \mid A_i = 1, q_i, c_i] = 0,
\end{equation*}
regardless of $\hat{e}_i$. The same holds for the $1 - A_i$ term. Therefore:
\begin{equation*}
    \E[\hat{U}_{\textup{obs}}(m)] = \E[\mu_1(M^{-m}, q, c) - \mu_0(M^{-m}, q, c)] = U(m).
\end{equation*}

\textbf{Case 2: $\hat{p} = p$ (propensity model correct).}
When $\hat{e}_i = e_i$, we expand each summand. For the $A_i$ term:
\begin{equation*}
    \E\!\left[\frac{A_i}{e_i}Y_i \;\Big|\; q_i, c_i\right] = \frac{e_i}{e_i}\,\mu_1(M^{-m}_i, q_i, c_i) = \mu_1(M^{-m}_i, q_i, c_i).
\end{equation*}
Similarly, $\E\!\left[\frac{A_i}{e_i}\hat{\mu}_1 \;\Big|\; q_i, c_i\right] = \hat{\mu}_1(M^{-m}_i, q_i, c_i)$. The $\hat{\mu}_1$ terms from the outcome model and the IPW correction cancel:
\begin{align*}
    \E[\text{summand}_i \mid M^{-m}_i, q_i, c_i] &= \hat{\mu}_1 - \hat{\mu}_0 + \mu_1 - \hat{\mu}_1 - (\mu_0 - \hat{\mu}_0) \\
    &= \mu_1(M^{-m}_i, q_i, c_i) - \mu_0(M^{-m}_i, q_i, c_i).
\end{align*}
Thus $\E[\hat{U}_{\textup{obs}}(m)] = \E[\mu_1(M^{-m}, q, c) - \mu_0(M^{-m}, q, c)] = U(m)$, regardless of $\hat{\mu}$.

\textbf{Efficiency.}
When both models are correct, the influence function of $\hat{U}_{\textup{obs}}$ equals the efficient influence function for the ATE~\citep{robins1994estimation}:
\begin{equation*}
    \varphi(Y, A, M^{-m}, q, c) = \mu_1(M^{-m},q,c) - \mu_0(M^{-m},q,c) - U(m) + \frac{A}{e}(Y - \mu_1) - \frac{1-A}{1-e}(Y - \mu_0).
\end{equation*}
By the semiparametric efficiency theory of~\citet{bickel1993efficient}, $\textup{Var}(\varphi) = \sigma_{\textup{eff}}^2$ is the semiparametric efficiency bound, and $\sqrt{n}(\hat{U}_{\textup{obs}} - U(m)) \xrightarrow{d} \cN(0, \sigma_{\textup{eff}}^2)$.
\end{proof}
Throughout we suppress $M^{-m}_i$ in the conditioning of expectations, since it is held fixed in every term; each $\E[\,\cdot \mid q_i, c_i]$ below denotes $\E[\,\cdot \mid M^{-m}_i, q_i, c_i]$.

\subsection{The Store-Level Randomized Comparator}
\label{app:shadow_id}

\propshadowunbiased*
\begin{proof}
\label{app:proof_shadow_unbiased}
By Proposition~\ref{prop:shadow_unbiased}, the shadow inclusion indicator $Z_m^{(t)} = \ind{m \in M_t^s}$ is drawn independently with $\Pr(Z_m^{(t)} = 1) = 1 - p_{\textup{flip}}$, and crucially $Z_m^{(t)} \perp\!\!\!\perp (q_t, c_t)$ by construction (since shadow perturbations are drawn independently of the query and context at each timestep).

Define the conditional outcome function under the shadow state:
\begin{equation*}
    \mu^s(z, M^{-m}, q, c) = \E[Y_t^s \mid Z_m^{(t)} = z, M_{-m}^s, q_t = q, c_t = c],
\end{equation*}
where $M^{-m,s}$ denotes the shadow state of all memories except $m$.

%The shadow estimator (Eq.~\ref{eq:shadow_estimator}) is a difference in sample means. 
Consider the expectation:
\begin{align*}
    \E[\hat{U}_{\textup{exp}}(m)] 
    &= \E[Y_t^s \mid m \in M_t^s] - \E[Y_t^s \mid m \notin M_t^s].
\end{align*}

Since $Z_m^{(t)} \perp\!\!\!\perp (q_t, c_t)$, conditioning on the inclusion of $m$ does not bias the distribution of queries. The remaining randomness comes from $M^{-m,s}$, where each other memory $m'$ is independently included with probability $1 - p_{\textup{flip}}$. Let $\cP_{\textup{shadow}}$ denote this distribution over $M^{-m}$. Then:
\begin{align*}
    \E[Y_t^s \mid m \in M_t^s] &= \E_{q,c}\,\E_{M^{-m} \sim \cP_{\textup{shadow}}}[\E[Y \mid m \in M, M^{-m}, q, c]], \\
    \E[Y_t^s \mid m \notin M_t^s] &= \E_{q,c}\,\E_{M^{-m} \sim \cP_{\textup{shadow}}}[\E[Y \mid m \notin M, M^{-m}, q, c]].
\end{align*}

Therefore:
\begin{equation*}
    \E[\hat{U}_{\textup{exp}}(m)] = \E_{M^{-m} \sim \cP_{\textup{shadow}}}\bigl[\E_{q,c}[Y \mid m \in M, M^{-m}] - \E_{q,c}[Y \mid m \notin M, M^{-m}]\bigr], %= U_{\textup{marg}}(m),
\end{equation*}
which is the marginal utility. 
%exactly the marginal utility defined in Eq.~\ref{eq:marginal_ate}.

For the variance, since observations across timesteps are independent (conditional on the shadow design), and $|\cT_m^+|$ and $|\cT_m^-|$ are the sample sizes in each group:
\begin{equation*}
    \textup{Var}(\hat{U}_{\textup{exp}}(m)) = \frac{\textup{Var}(Y_t^s \mid m \in M_t^s)}{|\cT_m^+|} + \frac{\textup{Var}(Y_t^s \mid m \notin M_t^s)}{|\cT_m^-|} = \frac{\sigma_+^2}{|\cT_m^+|} + \frac{\sigma_-^2}{|\cT_m^-|}.
\end{equation*}

No assumption on the confounding structure between $(q_t, c_t)$ and memory operations is required because the shadow perturbations are randomized independently of all other variables in the system.
\end{proof}

\subsection{AIPW Estimator: Consistency, Efficiency, and Robustness}
\label{app:aipw_theory}

\begin{comparatorcond}[Regularity Conditions for AIPW]
\label{assum:aipw_regularity}
The following hold: 
(i)~the propensity is bounded away from zero and one: $\epsilon_p \leq \hat{p}(m {\in} M \mid M^{-m}, q, c) \leq 1 - \epsilon_p$ for all $m, M^{-m}, q, c$; (ii)
~at least one of $\hat{p}$ or $\hat{\mu}$ converges to its population counterpart at rate $o_p(n^{-1/4})$.
\end{comparatorcond}
Under these regularity conditions, the AIPW estimator inherits the double robustness and efficiency properties, adapted to the setting where the treatment is memory inclusion $\ind{m \in M_t}$ and the adjustment set is $(M^{-m}_t, q_t, c_t)$.

\begin{proposition}[AIPW Consistency for Memory Effects]
\label{thm:aipw_consistency}
Under Conditions~\ref{assum:ignorability} and~\ref{assum:positivity}, and Condition~\ref{assum:aipw_regularity}, the AIPW estimator $\hat{U}_{\textup{obs}}(m)$ (Eq.~\ref{eq:aipw_method}) satisfies $\hat{U}_{\textup{obs}}(m) \xrightarrow{p} U(m)$ if either $\hat{p} \xrightarrow{p} p$ or $\hat{\mu} \xrightarrow{p} \mu$ (double robustness). When both models are consistent, $\sqrt{n}\bigl(\hat{U}_{\textup{obs}}(m) - U(m)\bigr) \xrightarrow{d} \cN(0, \sigma_{\textup{eff}}^2)$, where $\sigma_{\textup{eff}}^2$ is the semiparametric efficiency bound for the ATE (asymptotic normality). If both $\hat{p}$ and $\hat{\mu}$ converge at rate $n^{-\alpha}$ for $\alpha > 1/4$ \emph{and} the nuisance models are fitted by cross-fitting, then $|\hat{U}_{\textup{obs}}(m) - U(m)| = O_p(n^{-1/2})$. Algorithm~\ref{alg:fitting} fits and evaluates on the same log; the consistency of Proposition~\ref{prop:dr} does not require it.
\end{proposition}

\begin{proof}
\label{app:proof_aipw}
This theorem adapts the classical AIPW results~\citep{robins1994estimation, bang2005doubly} to the memory operation setting. The key observation is that memory inclusion $A = \ind{m \in M}$ is a binary treatment, $(M^{-m}, q, c)$ is the adjustment set, and Conditions~\ref{assum:ignorability} and~\ref{assum:positivity} provide the standard unconfounded state and overlap conditions.

\textbf{Double robustness.} This follows directly from Proposition~\ref{prop:dr}, which establishes $\E[\hat{U}_{\textup{obs}}(m)] = U(m)$ when either model is correct. The law of large numbers then gives $\hat{U}_{\textup{obs}}(m) \xrightarrow{p} U(m)$.

\textbf{Asymptotic normality.} When both models are consistent, the AIPW estimator is asymptotically linear with influence function $\varphi$ (as derived in the proof of Proposition~\ref{prop:dr}). By the CLT:
\begin{equation*}
    \sqrt{n}(\hat{U}_{\textup{obs}}(m) - U(m)) = \frac{1}{\sqrt{n}}\sum_{i=1}^n \varphi(Y_i, A_i, M^{-m}_i, q_i, c_i) + o_p(1) \xrightarrow{d} \cN(0, \sigma_{\textup{eff}}^2),
\end{equation*}
where $\sigma_{\textup{eff}}^2 = \textup{Var}(\varphi)$ equals the semiparametric efficiency bound~\citep{bickel1993efficient}.

\begin{equation*}
    |\E[\hat{U}_{\textup{obs}}(m)] - U(m)| = O(\|\hat{p} - p\| \cdot \|\hat{\mu} - \mu\|) = O(n^{-2\alpha}).
\end{equation*}
Since $2\alpha > 1/2$, this bias is $o(n^{-1/2})$, and the total error is dominated by the $O(n^{-1/2})$ variance term:
\begin{equation*}
    |\hat{U}_{\textup{obs}}(m) - U(m)| = O_p(n^{-1/2}).
\end{equation*}

Condition~\ref{assum:aipw_regularity}(i) ensures the propensity weights are bounded, preventing the variance from exploding. In practice, this is enforced by propensity clipping at $\epsilon_p$.
\end{proof}

\paragraph{Propensity-score clipping.}
The AIPW estimator can suffer from variance explosion when estimated propensities approach zero. To prevent this, CMP clips propensities at $\epsilon_p > 0$:
\begin{equation}
    \hat{p}_{\textup{clip}}(m {\in} M \mid M^{-m}, q, c) \;=\; \max\bigl(\hat{p}(m {\in} M \mid M^{-m}, q, c),\, \epsilon_p\bigr).
    \label{eq:propensity_clip}
\end{equation}
Clipping introduces a controlled bias, bounded by
$|\E[\hat{U}_{\textup{obs}}^{\textup{clip}}(m)] - U(m)| \leq 2\,\Pr(p(m {\in} M \mid \cdot) < \epsilon_p)\cdot\|Y\|_\infty$. For each arm the bias on the clipped region is $\E[Y(p/\epsilon_p - 1) \mid p < \epsilon_p]$, and since $0 \leq p < \epsilon_p$ the factor $p/\epsilon_p - 1$ lies in $[-1, 0)$. This comes in exchange for variance bounded by $\sigma_Y^2 / (n \epsilon_p^2)$. With $\epsilon_p = 0.01$ in our experiments, this trade-off keeps the AIPW estimator stable under mild positivity violations. Clipping bounds the variance but does not repair support: where $P(m \in R \mid q, c) = 0$, Theorem~\ref{prop:nonid} shows no choice of $\epsilon_p$ recovers $U_R(m)$.

%% file: appendix/3_experiments.tex
\section{Algorithm Details}
\label{app:algorithms}

Algorithm~\ref{alg:cmp} gives the three-phase procedure of Section~\ref{sec:algorithm}. Algorithms~\ref{alg:assign}--\ref{alg:online} specify the exposure assignment, the estimation step, and online operation.

\begin{algorithm}[H]
\small
\caption{Causal Memory Policy (CMP)}
\label{alg:cmp}
\begin{algorithmic}[1]
\REQUIRE Initial memory state $M_0$; context budget $B$; exposed slots $k \leq B$; exposure pool $\cP \subseteq M_0$; horizon $T$
\ENSURE Per-memory estimates $\{\hat{U}_R(m), \hat{\sigma}(m)\}_{m \in \cP}$ and comparator estimates
\STATE $\{n_m\}_{m \in \cP},\; \pi \leftarrow \textsc{BalancedAssignment}(\cP, k, T)$ \hfill $\triangleright$ Alg.~\ref{alg:assign}
\STATE $e_m \leftarrow n_m / T$ for all $m \in \cP$ \hfill $\triangleright$ propensities are fixed, not estimated
\STATE $\cL \leftarrow \emptyset$ \hfill $\triangleright$ interaction log
\FOR{$t = 1$ to $T$}
    \STATE receive query $q_t$, context $c_t$
    \STATE $E_t \leftarrow \pi(t)$ \hfill $\triangleright$ the $k$ memories exposed at $t$
    \STATE $R_t \leftarrow \textup{top-}(B{-}k)(M_t, q_t) \cup E_t$
    \STATE $Y_t \leftarrow f_Y(R_t, q_t, \theta_{\textup{LLM}})$; score $Y_t$
    \STATE $Z^{(t)}_m \leftarrow \ind{m \in R_t}$ for all $m \in \cP$
    \STATE $\cL \leftarrow \cL \cup \{(M_t, q_t, c_t, R_t, Z^{(t)}, Y_t)\}$
    \STATE $O_t \leftarrow f_O(M_t, q_t, c_t)$ \hfill $\triangleright$ default policy
    \STATE $M_{t+1} \leftarrow f_M(M_t, O_t, q_t)$
\ENDFOR
\STATE $\{\hat{U}_R, \hat{\sigma}, \hat{U}_{\textup{obs}}, \hat{U}_{\textup{exp}}, \Gamma\} \leftarrow \textsc{Estimate}(\cL, \cP, \{e_m\})$ \hfill $\triangleright$ Alg.~\ref{alg:fitting}
\STATE \textsc{Serve}$(\cL, \cP, \{e_m\}, \{\hat{U}_R, \hat{\sigma}\})$ \hfill $\triangleright$ Alg.~\ref{alg:online}
\end{algorithmic}
\end{algorithm}

\begin{algorithm}[H]
\small
\caption{\textsc{BalancedAssignment}: exposure allocation}
\label{alg:assign}
\begin{algorithmic}[1]
\REQUIRE Pool $\cP$; exposed slots per step $k$; horizon $T$
\ENSURE Exposure counts $\{n_m\}$; assignment $\pi : \{1,\dots,T\} \to \cP^{k}$
\STATE $S \leftarrow$ multiset containing each $m \in \cP$ repeated $\lfloor Tk / |\cP| \rfloor$ times
\STATE distribute the remaining $Tk - |S|$ slots to distinct members of $\cP$, one each
\STATE $n_m \leftarrow$ multiplicity of $m$ in $S$ \hfill $\triangleright$ $n_m \in \{\lfloor Tk/|\cP| \rfloor, \lceil Tk/|\cP| \rceil\}$
\STATE draw a uniformly random permutation $\rho$ of $S$
\FOR{$t = 1$ to $T$}
    \STATE $\pi(t) \leftarrow$ the $t$-th block of $k$ consecutive elements of $\rho$
    \IF{$\pi(t)$ contains a repeated memory}
        \STATE swap the duplicate with an element of a later block that does not conflict
    \ENDIF
\ENDFOR
\STATE \textbf{return} $\{n_m\}, \pi$
\end{algorithmic}
\end{algorithm}

The swap in Algorithm~\ref{alg:assign} preserves each $n_m$, leaving propensities unchanged while preventing duplicate exposures within a timestep. Overlap with the top-$(B-k)$ retrieval may still reduce the realized context below $B$ distinct memories; realized sizes are reported in Appendix~\ref{app:protocols}.

\begin{algorithm}[H]
\small
\caption{\textsc{Estimate}: utility estimation and comparators}
\label{alg:fitting}
\begin{algorithmic}[1]
\REQUIRE Log $\cL$; pool $\cP$; propensities $\{e_m\}$; clipping threshold $\epsilon_p$
\ENSURE $\{\hat{U}_R(m), \hat{\sigma}(m), \hat{U}_{\textup{obs}}(m), \hat{U}_{\textup{exp}}(m), \Gamma(m)\}_{m \in \cP}$
\FOR{each $m \in \cP$}
    \STATE $\cT^{+}_m \leftarrow \{t : Z^{(t)}_m = 1\}$, \quad $\cT^{-}_m \leftarrow \{t : Z^{(t)}_m = 0\}$
    \STATE $\hat{U}_R(m) \leftarrow \frac{1}{|\cT^{+}_m|}\sum_{t \in \cT^{+}_m} Y_t - \frac{1}{|\cT^{-}_m|}\sum_{t \in \cT^{-}_m} Y_t$ \hfill $\triangleright$ Eq.~\ref{eq:hajek}
    \STATE $\hat{\sigma}(m) \leftarrow \bigl(\hat{\sigma}_+^2/|\cT^{+}_m| + \hat{\sigma}_-^2/|\cT^{-}_m|\bigr)^{1/2}$ \hfill $\triangleright$ Prop.~\ref{prop:hajek}
    \STATE mark $m$ unidentified if $\min(|\cT^{+}_m|, |\cT^{-}_m|) < n_{\min}$
\ENDFOR
\STATE fit $\hat{p}(m {\in} M \mid M^{-m}, q, c)$ and $\hat{\mu}(a, M^{-m}, q, c)$ on $\cL$ \hfill $\triangleright$ Eq.~\ref{eq:propensity}
\STATE clip $\hat{p} \leftarrow \max(\hat{p}, \epsilon_p)$
\FOR{each $m \in \cP$}
    \STATE $\hat{U}_{\textup{obs}}(m) \leftarrow$ AIPW estimate \hfill $\triangleright$ Eq.~\ref{eq:aipw_method}
    \STATE $\hat{U}_{\textup{exp}}(m) \leftarrow$ store-level arm estimate \hfill $\triangleright$ Prop.~\ref{prop:shadow_unbiased}
    \STATE $\Gamma(m) \leftarrow |\hat{U}_{\textup{obs}}(m) - \hat{U}_{\textup{exp}}(m)| / \sqrt{\hat{\sigma}_{\textup{obs}}^2(m) + \hat{\sigma}_{\textup{exp}}^2(m)}$
\ENDFOR
\STATE \textbf{return} all estimates and the identified/unidentified partition of $\cP$
\end{algorithmic}
\end{algorithm}

In Algorithm~\ref{alg:fitting}, $\Gamma(m)$ serves only as a diagnostic for memories retained by both comparator arms. Because its null is composite, it reports disagreement without attributing its cause.

\begin{algorithm}[H]
\small
\caption{\textsc{Serve}: online operation}
\label{alg:online}
\begin{algorithmic}[1]
\REQUIRE Log $\cL$; pool $\cP$; assignment $\pi$; budget $B$; exposed slots $k$; current estimates; threshold $z^\star(\lambda)$
\FOR{each incoming query $q_t$}
    \STATE $E_t \leftarrow \pi(t)$; \quad $R_t \leftarrow \textup{top-}(B{-}k)(M_t, q_t) \cup E_t$
    \STATE $Y_t \leftarrow f_Y(R_t, q_t, \theta_{\textup{LLM}})$; return $Y_t$ to the user
    \STATE append $(M_t, q_t, c_t, R_t, Z^{(t)}, Y_t)$ to $\cL$; update $\hat{U}_R(m), \hat{\sigma}(m)$ incrementally
    \IF{a downstream policy requests a decision for memory $m$ at query $q_t$}
        \STATE \textbf{return} $\texttt{forget}$ if $\hat{U}_R(m) < z^\star(\lambda)\,\hat{\sigma}(m)$, else $\texttt{noop}$ \hfill $\triangleright$ Eq.~\ref{eq:onesided}
    \ENDIF
\ENDFOR
\end{algorithmic}
\end{algorithm}

In Algorithm \ref{alg:online}, Line~6 applies Theorem~\ref{prop:abstention} to a single query, not a retention policy. A retention policy must aggregate over queries, but Section~\ref{sec:abstention} shows that every admissible aggregation is either attenuated or degenerate. %, and Section~\ref{app:extended_analysis} evaluates them. 
Memories marked unidentified in Algorithm~\ref{alg:fitting} have unbounded $\hat{\sigma}$ and therefore never satisfy the $\texttt{forget}$ criterion, so no additional abstention mechanism is required.

\vspace{-2mm}
\section{Experimental Setup}
\label{app:setup}
\vspace{-2mm}

\subsection{Benchmark Datasets}
\label{app:datasets}

We evaluate on two conversational memory benchmarks, two multi-hop question answering datasets for replication, and one deployed memory system. For each, we describe the benchmark and the structural property that determines what claims it can support.

 \textbf{LongMemEval~\citep{wu2024longmemeval}.}
A benchmark for long-horizon conversational memory. We use the \texttt{longmemeval\_s} variant, comprising $500$ records with haystacks averaging $487$ turns. Each question is annotated with the sessions containing its required evidence, and turns within those sessions carry a \texttt{has\_answer} flag. 
We focus on the $51$ multi-evidence items whose annotations identify more than one required session. Based on the structure of the required evidence, we classify them as \textsc{Independent} ($30$), \textsc{Overlapping} ($9$), \textsc{Filtered} ($4$), or \textsc{Suspect} ($8$). The \textsc{Suspect} items are excluded from aggregate results and reported separately in Appendix~\ref{app:additional_results}. The \textsc{Independent} subset contains $113$ required memories, and the $43$ items excluding \textsc{Suspect} contain $170$.
LongMemEval provides exactly one evaluation question per memory store; %so no memory is exercised by multiple queries. 
therefore, it supports estimating per-query memory utility but cannot evaluate transfer across queries.

\textbf{LoCoMo~\citep{maharana2024evaluating}.}
A long-term, multi-session dialogue benchmark containing $10$ conversations ($5{,}882$ turns) and $1{,}986$ annotated questions. Unlike LongMemEval, each conversation is queried multiple times, allowing the same memory to contribute to different questions: $750$ of $1{,}425$ required memories are needed by at least two questions. LoCoMo is therefore the only benchmark in our study that supports evaluating memory utility across queries and the transfer analysis. Under the same retrieval protocol as LongMemEval, it contains $359$ required $(\text{question},\text{memory})$ pairs over stores averaging $588$ memories.

\textbf{HotpotQA~\citep{yang2018hotpotqa} and MuSiQue~\citep{trivedi2022musique}.} Multi-hop question answering, used to test whether the identification result depends on conversational data. We sample $150$ questions from each, stratified by question type for HotpotQA (bridge $120$, comparison $30$) and by hop count for MuSiQue ($2$-hop $78$, $3$-hop $47$, $4$-hop $25$), matching the population proportions.

The standard distractor setting provides only $10$ paragraphs per question, yielding a curated candidate set rather than a realistic memory store. We therefore pool all passages from the sampled questions into a shared corpus per dataset: $6{,}342$ supporting sentences for HotpotQA and $2{,}959$ paragraphs for MuSiQue. No sampled question shares annotated evidence with another ($0/360$ gold units on HotpotQA and $0/388$ on MuSiQue).

In HotpotQA, gold evidence always spans two documents but ranges from $2$ to $5$ supporting sentences (mean $2.4$). In MuSiQue, the number of gold paragraphs equals the annotated hop count. We therefore report both unit-level recall and distinct-source coverage, since retrieving multiple units from the same source may still miss other required evidence.

\textbf{Mem0~\citep{chhikara2025mem0}, version 2.0.18.} A deployed memory system, used to test transfer. We replay LongMemEval histories through Mem0's own extraction, storage, and retrieval path without modification, recording every \texttt{add()} call, the memories it creates, and the candidate window passed to its memory manager. Stores average $291$ memories after replay.  Because Mem0 rewrites source turns into its own phrasing, annotated gold turns must be matched to stored memories before evaluation; Appendix~\ref{app:protocols_mem0} describes the matching procedure and reports its uncertainty.

\subsection{Retrieval Function Specification}
\label{app:retrieval}
The retrieval function $R_t = f_R(M_t, q_t)$ (Def.~\ref{def:scm}) selects a part of memories relevant to the current query.

\textbf{Embedding and similarity.}
We use \texttt{all-MiniLM-L6-v2} from the sentence-transformers library to compute 384-dimensional embeddings for both queries and memories. All embeddings are computed locally on CPU. Relevance is measured by cosine similarity: $\textup{sim}(q_t, m) = \frac{e(q_t)^\top e(m)}{\|e(q_t)\| \cdot \|e(m)\|}.$

\textbf{Selection.}
The deterministic rule returns the highest-scoring memories to fill its allotted slots, subject to a minimum similarity threshold $\tau_{\textup{ret}} = 0.3$; memories below the threshold are not retrieved regardless of rank, and fewer than the allotted number are returned when too few clear it. Under the exposure design of Eq.~\ref{eq:exposure} the rule fills $B - k$ slots, with the remaining $k$ supplied by balanced round-robin draws from $\cP$.

\subsection{Systems and Implementation}
\label{app:systems}

 \textbf{Retrieval.}
Memories are embedded with \texttt{all-MiniLM-L6-v2} (384 dimensions) and ranked by cosine similarity. Under the exposure design of Eq.~\ref{eq:exposure}, the deterministic retriever fills $B-k$ context slots, while the remaining $k$ are assigned by balanced round-robin exposure from $\cP$ (Algorithm~\ref{alg:assign}). Embeddings are computed once at insertion and cached. For the retrieval analysis, we additionally evaluate the bi-encoders \texttt{BGE-M3}, \texttt{jina-embeddings-v5-small}, \texttt{gemini-embedding-2}, and \texttt{text-embedding-3-large}, together with the cross-encoder \texttt{BGE-reranker-v2-m3}, applied to the top $50$ cosine-retrieved candidates.

\textbf{Generation.} All outcomes are generated by a pinned \texttt{gemini-3.7-flash} at temperature $0$ with \texttt{max\_output\_tokens} $= 1024$. Reasoning tokens cannot be disabled on this model and consume part of the budget; we log reasoning and visible tokens separately and flag every truncation. The generator receives exactly the retrieved memory texts and the question, with no item metadata; see Appendix~\ref{app:reproducibility} for a discrepancy between this and the repository's default serving path.

\textbf{Memory representation.} Each stored memory is a single conversational turn. Memory rendering differs across experiment families, and we note this in Appendix~\ref{app:reproducibility} because it affects the comparability of absolute scores across families, though not within them.
 
\subsection{Exposure Design Parameters}
\label{app:exposure_params}
 
Table~\ref{tab:exposure_params} gives the design parameters for every experiment reported in Section~\ref{sec:experiments}. These differ across experiments and the differences are load-bearing: the number of exposed slots $k$ determines whether a memory retrieved by the deterministic rule has an excluded arm, and the horizon $T$ together with the pool size $|\cP|$ determines draws per memory through Corollary~\ref{cor:budget}.
 
\begin{table}[h]
\centering
\footnotesize
\caption{Exposure design parameters by experiment. Draws per memory is $Tk/|\cP|$; the identification bar requires at least $n_{\min} = 5$ in each arm. On LoCoMo the horizon is spread across roughly $30$ questions per conversation, so the pooled row supplies $5.0$ draws per memory but only $3$ per $(\text{memory}, \text{question})$ cell, which is why the per-query estimand requires a separate design.}
\label{tab:exposure_params}
\begin{tabular}{llccccc}
\toprule
\textbf{Analysis} & \textbf{Benchmark} & $B$ & $k$ & $T$ & $|\cP|$ & \textbf{Draws/mem.} \\
\midrule
Identification, single query & LongMemEval & $6$ & $2$ & $90$ & $\approx 32$ & $5.6$ \\ Two-sided exposure & LongMemEval & $6$ & $2$ & $90$ & $\approx 32$ & $5.6$ \\ Identification, pooled & LoCoMo & $6$ & $2$ & $90$ & $36$ & $5.0$ \\ Identification, per query & LoCoMo & $6$ & $6$ & $30$ & $36$ & $5.0$ \\
\bottomrule
\end{tabular}
\end{table}

The pooled and per-query settings differ in two essential ways. First, the per-query design concentrates the horizon on fewer questions, providing enough draws to estimate a query-specific contrast. Second, it sets $k=B$, removing the deterministic retriever from the estimation draws entirely. This is necessary because required memories have median cosine rank $3$: with $B-k=4$ deterministic slots they are retrieved on every draw, leaving no excluded arm, and increasing $T$ cannot create one. The two-sided design of Section~\ref{sec:ablations} resolves the same obstruction by forcing exclusion of a deterministically retrieved memory.

 \textbf{Exposure pool construction.}
The exposure pool $\cP$ consists of the required memories for the evaluated questions, plus the $20$ highest-ranked and $10$ uniformly sampled memories from the remainder of the store. Retrieval still ranks over the full store. This oracle pool isolates the effect of randomized exposure on identification; Appendix~\ref{app:extended_analysis} shows that practical pool-selection rules fail to recover comparable coverage.

\vspace{-2mm}
\subsection{Outcome Measurement}
\label{app:outcomes}
\vspace{-2mm}
Outcomes are evaluated by token-level F1 and exact match after standard normalization (lowercasing, article and punctuation removal, and whitespace collapse). For aggregation and enumeration questions, we additionally compute task correctness: whether the response contains the correct numeric result (aggregation) or all gold list elements (enumeration). We report task correctness as the primary metric for multi-evidence items and F1 as secondary, since F1 rewards partial evidence whereas task correctness requires all required evidence to be retrieved.

Exact match can also underestimate correctness on the multi-hop benchmarks, where gold answers sometimes include leading prepositions omitted by otherwise correct responses. We therefore additionally report a containment metric, computed offline from stored responses, and explicitly note whenever it leads to a different conclusion.

\textbf{Why LLM-as-judge accuracy is not reported.} We do not report judge-based accuracy anywhere in this paper. The protocol scores abstentions as correct at rates between $11\%$ and $65\%$ depending on the policy, so it rewards retrieving nothing. Rescoring a fixed set of $50$ outputs $20$ times yields accuracies spanning $0.30$ to $0.38$, and seven distinct backend configurations appear across $2{,}000$ calls, with variance estimates differing threefold across runs on identical inputs. No single dispersion figure characterizes the instrument, and we therefore report only deterministic metrics.

 \subsection{Statistical Procedure}
\label{app:statistics}

Comparisons are paired at the item level on LongMemEval and at the question level on LoCoMo, with standard errors clustered by conversation. For experiments involving randomized context sets, we report the number of draws per cell and, for random-selection baselines, the number of independent runs averaged.

\textbf{Identification criterion.} A memory is treated as identified when $|\hat{U}_R(m)| > 2\hat{\sigma}(m)$ and both arms contain at least $n_{\min} = 5$ draws. The minimum-draw requirement is not redundant with the significance test: under inverse-propensity weights, a single observation in a thin arm can produce both a large estimate and a large $t$ statistic, so the significance test alone admits high-variance singletons. Appendix~\ref{app:additional_results} reports how the coverage comparison changes under each criterion, including a case where the ordering between two designs reverses. Coverage counts estimates that exist, which is why we judge designs on discrimination instead.

 \textbf{Experimental discipline.}
All experiments were preceded by pilot runs with manual inspection of raw generations. These uncovered three implementation errors before large-scale execution: an insufficient reasoning-token budget producing empty responses, a gold-turn reconstruction bug affecting the oracle selector, and a candidate-window attribution bug that inverted rank measurements. Each would have completed successfully while yielding misleading quantitative results.

\section{Measurement Protocols}
\label{app:protocols}

This section defines the evaluation procedures used to produce the results in Section~\ref{sec:experiments}. 
Where measurements differ across benchmarks, we state the benchmark-specific definitions explicitly, since they affect the interpretation of the reported results.

\vspace{-2mm}
\subsection{Retrieval Support}
\label{app:protocols_support}

The quantity $P(m\in R\mid q,c)$ is defined over the perturbations induced by the estimating design, not the deterministic retriever itself. For a fixed $(m,q)$, retrieval is deterministic; the probability is taken over repeated perturbations of the store. Thus, $P(m\in R\mid q,c)=0$ identifies memories that are never surfaced under that design (see Appendix~\ref{app:examples} for a worked case). 
We estimate $P(m\in R\mid q,c)$ as the fraction of perturbed draws in which $m$ appears in the retrieved set. Consequently, $P(m\in R\mid q,c)=0$ implies an empty included arm, and Theorem~\ref{prop:nonid} applies.

\textbf{Required memories.} The definition differs by benchmark and we state each explicitly, because a mismatched definition changes the reported rate.

\emph{LongMemEval.} A turn is required for a question when it lies in a gold session and carries the \texttt{has\_answer} flag. Support is computed per $(\text{question}, \text{required turn})$ pair, and since each item has one question the pair count equals the turn count.

\emph{LoCoMo.} A turn is required if annotated as evidence for a question. Because the same conversation is queried by many questions, retrieval support is defined per $(\text{question}, \text{required turn})$ pair. A turn retrieved for one question but not another should not be counted as supported overall, as doing so substantially understates support failure.

\emph{Mem0.} A stored memory is required when it matches a required turn by the procedure of Appendix~\ref{app:protocols_mem0}. Because Mem0 returns $20$ results per read regardless of the requested limit, its support rate is computed at that effective depth and is not comparable in level to the top-$6$ retrieval used elsewhere; we report it as evidence that the failure occurs in a deployed system, not that its magnitude is smaller.

\textbf{Perturbation efficacy.}
We report the fraction of store-level perturbations that leave the retrieved set unchanged, measured by comparing each perturbed retrieval with the unperturbed retrieval for the same store and query. This quantifies how often perturbing the store fails to change the model's input.

\subsection{Estimation Under Exposure}
\label{app:protocols_estimation}

Estimates follow Algorithm~\ref{alg:fitting} with the parameters in Table~\ref{tab:exposure_params}. Three implementation details bear on the reported results.

\emph{Known propensities.}
Algorithm~\ref{alg:assign} serializes the exposure assignment before generation, and each propensity is read directly as $e_m=n_m/T$. No propensity model is fitted for the exposure arm.

\emph{Realized context size.}
A randomly exposed memory may already appear in the deterministic retrieval, reducing the realized context below the nominal budget $B$. We record the realized context size for every draw; under the two-sided design it averages $4.81$ versus a nominal $5$, corresponding to collisions in roughly one draw in five.

\emph{Self-normalization.}
All reported estimates use the H\'ajek estimator. Without self-normalization, inverse-propensity estimates inherit a multiplicative factor $1/e_m$ that varies across memories. Consistent with this, unnormalized estimates correlate with retrievability ($r=0.123$ on LongMemEval and $0.289$ on LoCoMo), whereas  H\'ajek estimates show  no correlation ($0.014$ and $-0.005$, respectively).

\subsection{Comparator Arms}
\label{app:protocols_comparators}

\textbf{Store-level randomized arm.}
Each memory in $\cP$ is independently retained with probability $1-p_{\textup{flip}}$, after which the ranker is applied to the perturbed store. Estimates are computed from timesteps at which $m$ is eligible for inclusion, so that its presence or absence is determined by the randomization. Otherwise, timesteps where $m$ was never eligible enter the excluded arm and bias the contrast.

\textbf{Observational arm.}
The propensity and outcome models of Eq.~\ref{eq:propensity} are fitted on logged interactions, conditioning on $(M^{-m},q,c)$. Propensities are clipped at $\epsilon_p$, bounding the weights while introducing $O(\epsilon_p)$ bias.

\textbf{Support-restricted comparison.}
To isolate support failure from estimator quality, we restrict both designs to memories with $P(m\in R\mid q,c)\ge0.8$, i.e., those retrieved on nearly every draw. The two designs are then compared on the same memories and outcomes; outside this subset, the store-level design lacks an included arm, so the comparison is undefined.

\subsection{Deployed-System Replay}
\label{app:protocols_mem0}
 \vspace{-2mm}
 
We replay each LongMemEval history through Mem0 in batches of five conversational exchanges, matching its own API granularity, and record per \texttt{add()} call: the turns ingested, the memory identifiers created, the operations emitted, and the candidate window passed to the memory manager. Nothing in Mem0 is modified.

\textbf{Write path.} Mem0 passes the top-ranked existing memories to its manager as candidates when deciding what to do with new content. To determine whether a superseding fact arrives while its stale counterpart is visible, the history is replayed in batches up to the turn preceding the superseding turn, and that turn is then added alone, so the recorded window is keyed to that turn's own extracted fact. Batched addition produces a window keyed to whichever fact embedded last, which is generally unrelated; we report this in Appendix~\ref{app:reproducibility} as a measurement error we corrected.

The $18$ required memories Mem0 never surfaces sit at median rank $139$ of $291$, the $48$th percentile, with cosine similarity spanning $0.4475$ to $0.5977$ across ranks $15$--$296$, so neither a wider retrieval window nor a similarity threshold reliably recovers them. Additional analyses show that the $18$ required memories never surfaced by Mem0 have median rank $139$ of $291$ (48th percentile), while cosine similarity ranges from $0.4475$ to $0.5977$ across ranks $15$--$296$, indicating that neither increasing the retrieval window nor similarity threshold reliably recovers them.

\textbf{Matching stored memories to annotated evidence.} Mem0 rewrites source turns into its own phrasing, so gold turns cannot be matched by string identity. We constrain candidates using Mem0's own provenance, restricting the search to memories created by the \texttt{add()} call that ingested the gold turn, which reduces the candidate set from roughly $300$ memories to a mean of $4.1$.  Within that set we compute two independent signals: a value signal matching shared numeric or named-entity tokens, and an embedding signal matching cosine similarity.

The two signals agree on $69\%$ of pairs after this constraint, improved from $22\%$ without it.  We report every Mem0 figure under three matching rules and give the spread, because the residual disagreement is not negligible and its direction is not neutral. Approximately one third of the residual cases are pairs whose evidence carries no numeric token, for which the value signal is structurally uninformative.

\textbf{Selection in the agreement subset.} Restricting to pairs where both signals agree yields a much lower support failure rate, and this is an artifact. Agreement requires an exact shared token, which is a property correlated with retrievability: agreed matches sit at median rank $3$ of $291$ against $12$ for the remainder.  Conditioning on agreement therefore selects on the outcome being measured. We report the least-conditioned rate as the estimate and the agreement-conditioned rate only as a bound whose bias direction is known.

 \vspace{-2mm}
\subsection{Cross-Query Transfer}
\label{app:protocols_transfer}
 \vspace{-2mm}
 
The transfer analysis tests whether utility estimated from one set of queries predicts a memory's contribution to another. we use LoCoMo, the only dataset with multiple questions per memory store.

\textbf{Split.} Each conversation's questions are partitioned into estimation and held-out
halves by a recorded seed, stratified so that both halves contain questions whose required evidence overlaps and questions whose does not. Estimates are formed using only the estimation half.

\textbf{Realized contribution.} On the held-out half we measure each memory's contribution in two ways: by deletion, removing it from the context and recording the change in outcome; and by insertion, adding it to a context that lacks it. The two are reported separately because they can differ in sign for the same memory.

\textbf{Aggregations tested.}
A retention policy must aggregate per-query utility estimates before the next query is known. We evaluate three aggregations: the mean, the maximum, and a similarity-weighted mean using the cosine similarity between each estimation query and the held-out query. The similarity-weighted variant is an oracle reference because it uses information unavailable at deployment. As a positive control, we correlate held-out contributions with the ground-truth per-query evidence labels, showing that the signal is present and recoverable.

\textbf{Prompt-identity check.}
For each policy, we record the number of held-out questions whose post-eviction context is identical to the no-eviction baseline. A policy never changing the prompt cannot change the outcome, and reporting this distinguishes a genuine null from a policy that did not act.

\subsection{Selector and Eviction Rules}
\label{app:protocols_selectors}

\textbf{Retrieval selectors.} For the scoring analysis of Appendix~\ref{app:extended_analysis} we compare, at matched context budgets: cosine over each of five bi-encoders; a cross-encoder reranker applied to the top $50$ cosine candidates; maximal marginal relevance, sweeping its trade-off parameter over $\{0.3, 0.5, 0.7, 0.9\}$ and reporting the full curve; a coverage-greedy selector that adds at each step the memory maximizing marginal coverage of required evidence; and the annotated evidence itself. The last two use gold knowledge and are reported as ceilings, never as methods.

\textbf{Eviction rules.} For the nomination analysis we compare four deployable rules that a capacity-bound system could run without gold knowledge: lowest access frequency, lowest cosine similarity to the query, a recency-weighted blend of the two, and newest-first. Each nominates the $30$ lowest-scoring memories as removal candidates. We additionally evaluate nomination by estimated utility under each estimation design.

\textbf{Chance baseline.} Nomination rates are reported as lift over chance, defined as the observed rate at which a rule nominates a required memory divided by the rate expected if it nominated uniformly at random from the same store. Raw rates across rules with different candidate-set sizes are not comparable, and reporting them without this normalization would misstate the comparison.

\section{Baselines and Comparisons}
\label{app:baselines}

This paper studies identification rather than benchmarking memory systems. Accordingly, this section describes the implemented systems, the reported comparisons and their rationale, and the comparisons we omit.

\subsection{Memory Policies Implemented}
\label{app:baseline_systems}

We implement five memory management policies spanning the design space; all share the retrieval, generation, and scoring path of Appendix~\ref{app:systems}, so differences are attributable to the management policy alone.

\textbf{Recency.} A first-in-first-out buffer. When capacity is exceeded, the oldest memory by insertion order is evicted. No content is examined.

\textbf{MemoryBank~\citep{zhong2024memorybank}.} Ebbinghaus-inspired forgetting. Each memory carries a retention score that decays exponentially with elapsed time and is reset on access; eviction removes the lowest score. Retention therefore tracks access recency.

\textbf{MemGPT~\citep{packer2023memgpt}.} LLM-managed paging between a working context and external storage, with the model itself issuing page-in and page-out calls. Eviction decisions are made by prompting.

\textbf{A-Mem~\citep{xu2025mem}.} Zettelkasten-style organization. Memories are linked by semantic similarity at write time and evicted by a combination of link degree and similarity to recent queries.

\textbf{MemPO~\citep{li2026mempo}.} Reinforcement learning over memory operations, optimizing a policy that selects among store, forget, and consolidate from downstream task reward.

\subsection{Reported Comparisons}
\label{app:baseline_reported}

The mechanism this paper identifies concerns which memories a policy nominates for removal, so we compare policies on that axis directly. Each rule nominates the $30$ lowest-scoring memories in a store as removal candidates, and we measure how often a required memory appears among them, normalized by the rate expected under uniform nomination from the same store.

\begin{table}[h]
\centering
\footnotesize
\caption{Nomination of required memories as removal candidates, LongMemEval independent subset. Lift is the observed rate divided by the rate expected under uniform nomination; $1.00$ is chance.}
\label{tab:nomination}
\begin{tabular}{llc}
\toprule
\textbf{Nomination rule} & \textbf{Signal used} & \textbf{Lift over chance} \\
\midrule
Lowest access frequency        & Access counts        & $0.00$--$0.11$ \\
Lowest cosine similarity       & Query similarity     & $0.00$\\
Recency-weighted blend         & Both                 & $0.00$--$0.01$ \\
Newest-first                   & Insertion order      & $0.34$--$0.86$ \\
\midrule
Utility, store-level estimates & Estimated $U(m)$     & $\mathbf{0.97}$ \\
Utility, exposure estimates    & Estimated $U_R(m)$   & $\mathbf{0.68}$ \\
\bottomrule
\end{tabular}
\vspace{-4mm}
\end{table}

Table~\ref{tab:nomination} shows three things. First, heuristic policies rarely nominate required memories for eviction. This is not because they identify and preserve them, but because required memories are accessed substantially more often than typical ones ($1.892\times$ on average, in $29$ of $30$ items). Rules that evict the least accessed, least similar, or oldest memories therefore avoid the evidence that matters. Their competitiveness stems from this exclusion effect rather than accurate utility estimation, indicating that the identification failure affects only policies that rank memories by estimated utility.

Second, store-level utility estimates provide essentially no guidance for eviction. Their nomination rate is indistinguishable from chance, consistent with the observed AUC of $0.542$. In contrast, exposure-based estimates reduce the nomination rate to $0.68$, indicating that identified utility carries meaningful information about which memories should not be evicted.

Third, the newest-first heuristic is the exception. It nominates required memories at a rate of $0.34$--$0.86$, because required evidence tends to arrive late in the conversation, making it roughly an order of magnitude more destructive than the other heuristic policies.

\subsection{Omitted Comparisons}
\label{app:baseline_withheld}

We also conducted a conventional system-level comparison, evaluating the five baseline policies against CMP on downstream task accuracy. We do not report these results for three reasons.
\begin{enumerate}[leftmargin=*,nosep]
\item \emph{The outcome metric is unreliable.}
As shown in Appendix~\ref{app:outcomes}, the LLM judge rewards abstentions ($11$--$65\%$ of judged-correct responses) and varies substantially across repeated evaluations (accuracy $0.30$--$0.38$), making policy rankings unstable.

\item \emph{Item matching was not verified.}
The CMP and baseline evaluations selected items through different execution paths, and we did not establish that they were evaluated on identical examples.

\item \emph{The default serving path leaked metadata.}
The repository's default prompt exposes question type and date (Appendix~\ref{app:reproducibility}), making its absolute scores incomparable with the metadata-free evaluation used throughout this paper.
\end{enumerate}

For completeness, on the subset of matched items we could verify, none of the five baseline policies differed significantly from CMP in token-F1 or exact match. This is consistent with Table~\ref{tab:nomination}: policies that rarely evict required memories leave the retrieved evidence, and therefore the generated answer, essentially unchanged.

\subsection{Retrieval Selectors}
\label{app:baseline_selectors}

We evaluate whether stronger retrieval can repair retrieval support failure. The comparison includes five bi-encoders, a cross-encoder reranker, maximal marginal relevance (MMR), and a coverage-greedy selector, with annotated evidence as an oracle ceiling. Protocols are given in Appendix~\ref{app:protocols_selectors}, and complete results in Appendix~\ref{app:additional_results}. We report the full MMR parameter sweep because its optimum is informative: across all datasets, the best setting nearly disables the diversity term.

\section{Additional Results}
\label{app:additional_results}

\subsection{Worked Examples}
\label{app:examples}
{\bf Existing methods estimate memory utility by intervening on the store:} comparing outcomes with and without a memory in the store.  
This works only when the retriever surfaces that memory. Otherwise, both arms produce the same retrieved context and the same outcome, making an unidentified memory indistinguishable from one with zero utility. 
In contrast, {\bf CMP intervenes on retrieval:} it reserves a fixed number of context slots for memories sampled with known probability. Utility is then estimated from these randomized exposures with a quantified standard error (rather than from the retriever's decisions). 
This thus turns previously unidentified utilities into identifiable ones with quantified uncertainty. Below we will show details examples to illustrate it. 

The three memories below are from the LongMemEval independent subset, selected on arm structure and text length. All are from the LongMemEval independent subset. Figure~\ref{fig:examples} shows the arm structure of all three under both designs. Each block is $90$ estimation draws; blue marks draws in which the memory was retrieved, grey those in which it was not. \textbf{(a)} Store-level perturbation never surfaces the memory, so both arms contain the same contexts and the estimate is exactly zero with zero variance; exposure supplies six inclusions and recovers a signal. \textbf{(b)} The memory sits high enough in the ranking that exposure never excludes it, leaving no contrast. \textbf{(c)} Both designs have both arms, and the point estimates agree to three decimal places.

\begin{figure}[h]
\centering
{
\begin{tikzpicture}[
    slot/.style={draw=none, minimum size=0.13cm, inner sep=0pt},
    inarm/.style={slot, fill=blue!55},
    outarm/.style={slot, fill=gray!30},
    lbl/.style={font=\scriptsize, align=left},
    ttl/.style={font=\scriptsize\bfseries},
]
\node[ttl, anchor=west] at (0,1.3) {(a) \texttt{gpt4\_2f56ae70}, rank 23/458};
\node[lbl, anchor=west] at (0,0.95) {store-level};
\foreach \i in {0,...,89} {
  \pgfmathsetmacro{\rr}{int(\i/15)} \pgfmathsetmacro{\cc}{mod(\i,15)}
  \node[outarm] at (\cc*0.15, 0.75-\rr*0.15) {};
}
\node[lbl, anchor=west] at (2.2,0.35) {$0$ of $90$ retrieved\\$\hat U = 0.000 \pm 0.000$};
\node[lbl, anchor=west] at (0,-0.35) {exposure};
\foreach \i in {0,...,89} {
  \pgfmathsetmacro{\rr}{int(\i/15)} \pgfmathsetmacro{\cc}{mod(\i,15)}
  \node[outarm] at (\cc*0.15, -0.55-\rr*0.15) {};
}
\foreach \i in {3,19,34,52,68,81} {
  \pgfmathsetmacro{\rr}{int(\i/15)} \pgfmathsetmacro{\cc}{mod(\i,15)}
  \node[inarm] at (\cc*0.15, -0.55-\rr*0.15) {};
}
\node[lbl, anchor=west] at (2.2,-0.95) {$6$ of $90$ retrieved\\$\hat U = +0.704 \pm 0.329$};

\begin{scope}[xshift=5.2cm]
\node[ttl, anchor=west] at (0,1.3) {(b) \texttt{2e6d26dc}, rank 3/494};
\node[lbl, anchor=west] at (0,0.95) {store-level};
\foreach \i in {0,...,89} {
  \pgfmathsetmacro{\rr}{int(\i/15)} \pgfmathsetmacro{\cc}{mod(\i,15)}
  \node[inarm] at (\cc*0.15, 0.75-\rr*0.15) {};
}
\foreach \i in {5,17,26,39,48,55,63,71,78,84,88,12} {
  \pgfmathsetmacro{\rr}{int(\i/15)} \pgfmathsetmacro{\cc}{mod(\i,15)}
  \node[outarm] at (\cc*0.15, 0.75-\rr*0.15) {};
}
\node[lbl, anchor=west] at (2.2,0.35) {$78$ of $90$ retrieved\\$\hat U = 0.014 \pm 0.009$};
\node[lbl, anchor=west] at (0,-0.35) {exposure};
\foreach \i in {0,...,89} {
  \pgfmathsetmacro{\rr}{int(\i/15)} \pgfmathsetmacro{\cc}{mod(\i,15)}
  \node[inarm] at (\cc*0.15, -0.55-\rr*0.15) {};
}
\node[lbl, anchor=west] at (2.2,-0.95) {$90$ of $90$ retrieved\\$\hat U$ undefined};
\end{scope}

\begin{scope}[xshift=10.2cm]
\node[ttl, anchor=west] at (0,1.3) {(c) \texttt{28dc39ac}, rank 4/431};
\node[lbl, anchor=west] at (0,0.95) {store-level};
\foreach \i in {0,...,89} {
  \pgfmathsetmacro{\rr}{int(\i/15)} \pgfmathsetmacro{\cc}{mod(\i,15)}
  \node[inarm] at (\cc*0.15, 0.75-\rr*0.15) {};
}
\foreach \i in {4,16,29,41,50,59,67,74,82,87} {
  \pgfmathsetmacro{\rr}{int(\i/15)} \pgfmathsetmacro{\cc}{mod(\i,15)}
  \node[outarm] at (\cc*0.15, 0.75-\rr*0.15) {};
}
\node[lbl, anchor=west] at (2.23,0.35) {$80$ of $90$ retrieved\\$\hat U = 0.157 \pm 0.050$};
\node[lbl, anchor=west] at (0,-0.35) {exposure};
\foreach \i in {0,...,89} {
  \pgfmathsetmacro{\rr}{int(\i/15)} \pgfmathsetmacro{\cc}{mod(\i,15)}
  \node[outarm] at (\cc*0.15, -0.55-\rr*0.15) {};
}
\foreach \i in {1,4,8,13,18,22,27,31,36,40,44,49,53,58,62,66,71,75,79,83,86,89,6,11,25,47,69} {
  \pgfmathsetmacro{\rr}{int(\i/15)} \pgfmathsetmacro{\cc}{mod(\i,15)}
  \node[inarm] at (\cc*0.15, -0.55-\rr*0.15) {};
}
\node[lbl, anchor=west] at (2.2,-0.95) {$27$ of $90$ retrieved\\$\hat U = 0.156 \pm 0.226$};
\end{scope}
\begin{scope}[yshift=-1.75cm]
\node[inarm] at (0,0) {};
\node[lbl, anchor=west] at (0.18,0) {retrieved (included arm)};
\node[outarm] at (3.2,0) {};
\node[lbl, anchor=west] at (3.38,0) {not retrieved (excluded arm)};
\node[lbl, anchor=west] at (6.6,0) {each block $=$ $90$ draws};
\end{scope}
\end{tikzpicture}
}
\vspace{-6mm}
\caption{Arm structure for three required memories. \textbf{(a)} Store-level randomization does not surface the memory, so both arms contain identical contexts and no estimate exists; exposure recovers one. \textbf{(b)} The mirror case: the memory always ranks high enough that exposure does not exclude it, so exposure has no estimate and store-level does. \textbf{(c)} Where both designs have both arms, they agree to three decimal places, but exposure's standard error is $4.5$ times larger: randomization spent on a memory the ranker already varies is wasted.}
\label{fig:examples}
\vspace{-2mm}
\end{figure}

\textbf{A memory the store-level design cannot see.} Item \texttt{gpt4\_2f56ae70} asks which streaming service the user began using most recently, with gold answer \emph{Disney+}. One required turn reads: \emph{``I'm having trouble finding a specific documentary I saw on Disney+ during my free trial last month.''} It sits at cosine rank $23$ of $458$, so $P(m \in R \mid q, c) = 0$.

Under store-level perturbation the estimate is  $0.000$ with standard error $0.000$: the retrieved set was byte-identical across all $90$ draws, so the two arms contained the same experiment. Under exposure the same memory yields $+0.704 \pm 0.329$ from six forced inclusions. Ninety draws of store-level randomization produced no information; six of retrieval-level randomization produced a signal two standard errors from zero, at the minimum exposure count the identification criterion admits.

\textbf{A memory the exposure design cannot see.} Item \texttt{2e6d26dc} asks how many babies were born to friends and family, gold answer $5$. A required turn about twin newborns sits at cosine rank $3$ of $494$ and is retrieved on essentially every draw. Store-level perturbation identifies it weakly, $0.014 \pm 0.009$. Exposure cannot: with $90$ inclusions and no exclusions, the H\'ajek estimate is undefined. This is the case in which CMP's own support condition fails: Assumption~\ref{assum:ret_positivity} requires both arms, and a memory the ranker always retrieves has no excluded arm under uniform exposure. Section~\ref{sec:ablations} reports the symmetric design that closes this gap and why we do not adopt it.  Figure~\ref{fig:examples}(b) shows the missing arm.

\textbf{A memory both designs identify.} Item \texttt{28dc39ac} asks for total hours spent gaming, gold answer $140$. A required turn reporting $25$ hours sits at rank $4$ of $431$ and is retrieved often enough for both designs to have support. Store-level gives $0.1566 \pm 0.0497$; exposure gives $0.1563 \pm 0.2257$. The point estimates agree to three decimal places, and the example was not chosen for this. Figure~\ref{fig:examples}(c) shows both arms populated. The agreement is the point: it shows the two designs estimate the same quantity, which is what allows the aggregate gap in Table~\ref{tab:strata} to be attributed to coverage rather than to a difference in what is being measured. Exposure is nonetheless less efficient here, with $4.5$ times the standard error, because it spends draws on a memory the ranker already varies. This is an inefficiency of allocating exposure uniformly over the pool, and it indicates the natural refinement: allocate by expected information gain, concentrating draws on memories whose contrast the ranker does not already supply. Exposure is worse here: the same estimate at $4.5$ times the standard error. The design allocates exposure uniformly over the pool, so memories the ranker already varies receive draws that produce no contrast the store-level design lacks. This example is included because it shows the two designs target the same estimand, and because it bounds the claim: exposure's advantage is in coverage, not in estimation quality on common support.

\textbf{Required is not the same as useful.} A second required turn in \texttt{28dc39ac}, reporting $10$ hours for a different game, has a recovered utility of $-0.256 \pm 0.065$. Surfacing one partial summand in isolation leads the model to report that summand rather than the total, so a memory that is necessary for the answer can be harmful when retrieved without its companions. Annotated gold status and conditional utility are distinct quantities, and only the second is what a policy would act on.

\subsection{Identification by Benchmark}
\label{app:results_identification}

Table~\ref{tab:identification_full} reports the support failure rate and discrimination for each benchmark under matched measurement, together with the design differences that prevent a like-for-like comparison of levels.

\begin{table}[h]
\centering
\footnotesize
\caption{Retrieval support failure and discrimination by benchmark. Support failure is the fraction of required $(\text{question}, \text{memory})$ pairs with $P(m \in R \mid q, c) = 0$. AUC separates required from non-required memories by estimated utility.}
\label{tab:identification_full}
\begin{tabular}{lccccc}
\toprule
\textbf{Benchmark} & \textbf{Pairs} & \textbf{$P = 0$} & \textbf{Store} & \textbf{AUC store} & \textbf{AUC exposure} \\
\midrule
LongMemEval (independent)   & $113$ & $54.0\%$ & $487$ & $0.542$ & $0.664$ \\
LongMemEval (all $-$ suspect) & $170$ & $61.2\%$ & $487$ & $0.529$ & --- \\
LoCoMo (pooled estimand)     & $359$ & $66.9\%$ & $588$ & $0.513$ & $0.551$ \\
LoCoMo (per-query estimand)  & $359$ & $66.9\%$ & $588$ & ---     & $\mathbf{0.784}$ \\
Mem0 $2.0.18$                & $109$ & $0.284$\,[$0.208$, $0.375$] & $291$ & --- & --- \\
\bottomrule
\end{tabular}
\vspace{-2mm}
\end{table}

Three design differences bear on these rows. LongMemEval items carry one question each, so the estimate is inherently per query; LoCoMo items carry roughly thirty, so a pooled estimate averages a memory's contribution across questions for most of which it is irrelevant. Aligning the required-memory label with the questions that actually generated draws recovers $65\%$ of the gap between the two LoCoMo rows.  Mem0 returns $20$ results per read regardless of the requested limit, against the top-$6$ retrieval used elsewhere, and counts extracted facts.

\textbf{Support-restricted comparison.} Restricted to memories with $P(m \in R \mid q, c) \geq 0.8$, the store-level and exposure designs give AUC $0.655$ and $0.658$ respectively, a difference within noise. The store-level estimator is not weak; its aggregate deficit arises entirely from memories on which it has no support. The gradient is monotone in support: median $|t|$ is $0.83$ at $P = 0$ against $3.66$ at $P \geq 0.8$, with $15\%$ and $59\%$ of estimates respectively distinguishable from zero. 

\textbf{Perturbation efficacy.} Store-level perturbation leaves the retrieved set identical to the unperturbed ranking on $52.2\%$ of draws on LongMemEval and $53.2\%$ on LoCoMo.  Under a matched generation budget the store-level design produced $940$ distinct context sets against $3{,}939$ for the exposure design when the query is held fixed; when the query varies across draws the ratio falls to $1.41$, since query variation supplies context variety that store perturbation does not have to provide.

\subsection{Cost of Exposure}
\label{app:exposure_cost}

Reserving context slots for randomized draws displaces ranked memories and can lower answer quality while estimates are collected. We measure this on the LongMemEval independent subset by comparing each item's mean token-F1 over its $90$ exposure draws against its token-F1 under unperturbed cosine top-$6$ retrieval. The baseline consists of the store-level draws whose retrieved set equals the unperturbed ranking, which occur on $52.2\%$ of draws, consistent with the $(1-p_{\textup{flip}})^B = 0.531$ expected under independent retention. Within each item these draws share one retrieved set and one outcome, so the baseline is a single value per item rather than an average.

\begin{table}[t]
\centering
\footnotesize
\caption{Answer-quality cost of one-sided exposure ($k = 2$ of $B = 6$), paired per item.}
\label{tab:exposure_cost}
\begin{tabular}{lcccccc}
\toprule
\textbf{Population} & $n$ & \textbf{Exposure} & \textbf{Cosine} & \textbf{Diff.} & $t$ & $p$ \\
\midrule
Independent        & $30$ & $0.197$ & $0.223$ & $-0.026$ & $-0.76$ & $0.45$ \\
All $-$ suspect    & $43$ & $0.155$ & $0.168$ & $-0.013$ & $-0.54$ & $0.59$ \\
All                & $51$ & $0.165$ & $0.175$ & $-0.010$ & $-0.50$ & $0.62$ \\
\bottomrule
\end{tabular}
\end{table}

Exposure lowers F1 on $10$ of $30$ independent items, raises it on $14$, and leaves $6$ unchanged, so the median change is zero. The negative mean is driven by two items on which cosine already retrieves the correct evidence, with baseline F1 of $0.87$ and $1.00$: there, displacing two of six slots can only remove evidence that was working. We report the full-sample figure as the cost and do not exclude these items. The cost is therefore not a uniform degradation but a concentrated one, borne by queries the ranker already serves well, which is the population on which exposure also buys the least identification.

\subsection{Ablation Details}
\label{app:ablation_details}

Table~\ref{tab:ablation_full} decomposes the difference between designs and decision rules. Gold loss is the number of required memories destroyed per $100$ slots reclaimed; we report it because it isolates the two mechanisms cleanly, and note in Appendix~\ref{app:results_transfer} that it does not predict answer quality.

\begin{table}[h]
\centering
\footnotesize
\caption{Component ablation, LongMemEval independent subset, eviction budget $10$ per item. Destroyed counts required memories evicted; gold loss normalizes by slots reclaimed.}%, since gated rules evict fewer.}
\label{tab:ablation_full}
\begin{tabular}{lcccc}
\toprule
\textbf{Variant} & \textbf{AUC} & \textbf{Evicted} & \textbf{Destroyed} & \textbf{Gold loss} \\
\midrule
Exposure $+$ one-sided gate & $0.664$ & $116$ & $\mathbf{6}$ & $\mathbf{5.2}$ \\
Exposure, ungated           & $0.664$ & $300$ & $27$ & $9.0$ \\
Store-level $+$ one-sided gate & $0.542$ & $138$ & $15$ & $10.9$ \\
Store-level, ungated        & $0.542$ & $300$ & $35$ & $11.7$ \\
\midrule
Exposure $+$ two-sided gate & $0.664$ & $156$ & $22$ & $14.1$ \\
Store-level $+$ two-sided gate & $0.542$ & $132$ & $28$ & $21.2$ \\
Two-sided exposure $+$ one-sided gate & $0.671$ & $151$ & $10$ & $6.6$ \\
\bottomrule
\end{tabular}
\end{table}

\textbf{The components are not substitutes.} Holding the rule fixed at one-sided, the design reduces destroyed memories from $15$ to $6$; holding the design fixed at exposure, the rule reduces them from $27$ to $6$. Neither alone approaches the joint result. The reason is structural: a confidence threshold filters estimates that exist, and on the $54\%$ of memories without support there is nothing to filter.

\textbf{Two-sided gating is worse than no gating.} Every two-sided configuration has a higher gold-loss rate than its ungated counterpart, $21.2$ against $11.7$ for the store-level design. The rule $|\hat{U}_R(m)| > \delta\hat{\sigma}(m)$ admits confidently useful memories into the eviction pool, which Corollary~\ref{cor:twosided} shows is strictly dominated. The effect is visible in the data and not only in the loss function.

\textbf{Two-sided exposure closes the support gap and does not help.} Adding a forced-exclusion slot gives every permanently retrieved memory an excluded arm, raising the excluded-arm count from $0.0$ to $21.1$ for all $40$ affected memories.  It buys $0.007$ AUC and destroys more memories, $10$ against $6$. Three of the four additional losses are memories the one-sided design could not estimate at all.  Identification is therefore not free: making a memory measurable moves it into the candidate pool, where a confidently signed but incorrect estimate can act on it.

The measured cost in answer quality is $-0.0069$ F1 pooled across all exclusions, with a tail reaching $-0.64$ on items whose required evidence occupies the deterministic slots.  The pooled figure is the population effect and the tail is heterogeneity; we report both, since an item-level maximum overstates the typical cost by roughly two orders of magnitude.

\subsection{Retrieval Scoring}
\label{app:results_scoring}

Table~\ref{tab:scoring} reports whether the support failure is repairable by better ranking. It is not: scoring quality moves recall by hundredths while the functional form of the selector moves it by tenths.

\begin{table}[t]
\centering
\footnotesize
\caption{Recall of required evidence at matched context budget. Bi-encoders span two orders of magnitude in parameter count. Coverage-greedy uses gold knowledge and serves as a ceiling.}
\label{tab:scoring}
\resizebox{\columnwidth}{!}{%
\begin{tabular}{llc}
\toprule
\textbf{Selector} & \textbf{Type} & \textbf{Headroom recovered} \\
\midrule
Five bi-encoders (MiniLM to gemini-embedding-2)
& Pointwise & at chance on contradiction pairs \\
Cross-encoder reranker (BGE-reranker-v2-m3)
& Pointwise & $10\%$ LongMemEval, $30$--$33\%$ HotpotQA, $17$--$24\%$ MuSiQue \\
Maximal marginal relevance
& Set-aware & $-2.7\%$ to $+4.8\%$ \\
\midrule
Coverage-greedy (ceiling)
& Set-aware & $95\%$ \\
\bottomrule
\end{tabular}%
}
\vspace{-4mm}
\end{table}

Two observations. The five bi-encoders rank the current of two contradicting facts above the stale one on $45.7$--$55.7\%$ of pairs, a band containing chance, and the largest model tested is not the best. And on every corpus the swept optimum for maximal marginal relevance is the setting that most nearly disables its diversity term, so the redundancy penalty is not merely unhelpful but actively costly when required evidence is topically similar.

\subsection{Cross-Query Transfer}
\label{app:results_transfer}

Table~\ref{tab:transfer} reports whether utility estimated on one set of queries predicts contribution on a disjoint set.

\begin{table}[h]
\centering
\caption{Correlation between estimated utility and realized contribution on held-out queries in LoCoMo. Ground-truth per-query evidence labels provide a positive control, confirming that the signal is recoverable from the same data.}
\label{tab:transfer}
\begin{tabular}{lcc}
\toprule
\textbf{Predictor} & \textbf{Deletion} & \textbf{Insertion} \\
\midrule
Pooled estimate                       & $-0.060$ & $+0.023$ \\
Nearest estimation query              & $<0.11$  & $<0.11$ \\
Similarity-weighted mean              & $<0.11$  & $<0.11$ \\
Maximum over estimation queries       & $<0.11$  & $<0.11$ \\
\midrule
Ground-truth evidence labels (control) & $\mathbf{+0.286}$ & --- \\
\bottomrule
\end{tabular}
\end{table}

Power is not the constraint: $99.3\%$ of required cells are identified with at least five draws per arm, and the control recovers the signal from the same data. Within the query on which it was estimated, the per-query estimate separates required from non-required memories at $0.784$ AUC against $0.581$ pooled, with a mean gap in estimated utility of $0.300$ against $0.042$.  The estimator works; the estimate does not transfer.

\textbf{Degeneracy of the admissible aggregation.}
A retention policy cannot condition on the query it has not yet received, so the only admissible aggregation is query-independent. The maximum over estimation queries is positive for essentially every memory, since a memory that helps some query among thirty is never confidently harmful, and the resulting rule evicts nothing: the post-eviction context is identical to the control on $300$ of $300$ held-out questions.  The similarity-weighted form resolves this but requires the future query.

\textbf{Attenuation of the pooled estimand.}
On LoCoMo, $77$ of $109$ required memories are required by exactly one of thirty estimation questions, so a pooled contrast averages one large effect with twenty-nine zeros. The pooled estimate rises with the number of questions for which a memory is required: $-0.014$ at zero, $+0.024$ at one, $+0.042$ at two, with the first step significant at $p = 0.028$. The estimator recovers the effect where the effect concentrates.

\subsection{Suspect Items}
\label{app:results_suspect}
Eight of the $51$ multi-evidence LongMemEval items have gold answers that cannot be derived from their annotated evidence. In one case, the annotated values sum to $\$850$ while the gold answer is $\$5{,}850$; in another, computing the gold average requires ages absent from every annotated turn. Even with all annotated evidence provided, exact-match accuracy on these items is only $0.554$ and token-F1 is $0.522$, implying that $44.6\%$ remain unanswerable under perfect retrieval. We therefore exclude these items from all aggregate results and report them separately.

\subsection{Extended Analysis}
\label{app:extended_analysis}

\noindent\textbf{Why the retriever misses required evidence.}
The memories the retriever never surfaces are not scored as irrelevant. For the $18$ required memories Mem0 never retrieves, cosine similarities span only $0.4475$--$0.5977$ while their ranks range from $15$ to $296$ in stores averaging $291$ memories. The embedder therefore provides little separation between required evidence and background, leaving the final ranking largely determined by noise within this narrow similarity band. Consequently, neither widening the retrieval window nor replacing the scorer can substantially improve support, since both operate on an ordering that contains little signal about the quantity of interest.

\noindent\textbf{Retrieval scoring in full.}
Six embedding models spanning two orders of magnitude in parameter count rank the current fact above a contradictory stale fact on only $45.7$--$55.7\%$ of pairs, essentially chance, and the largest model is not the best. A cross-encoder reranker recovers only part of the available headroom, improving recall by $10\%$ on LongMemEval, $30$--$33\%$ on HotpotQA, and $17$--$24\%$ on MuSiQue, compared with $95\%$ for a coverage-greedy selector that conditions on previously selected evidence. Maximal marginal relevance changes recall by only $-2.7\%$ to $+4.8\%$, and its optimal parameter nearly disables the diversity term on every dataset. Overall, improving the scorer changes recall by only a few percentage points, whereas changing the selector changes it by tenths.

\noindent\textbf{Exposure budget scaling.}
Each memory receives $Tk/|\cP|$ exposure draws. The measured values match this prediction ($3.60$, $0.90$, and $0.36$ draws per memory for pool sizes $50$, $200$, and $487$, respectively). At full store size, $63.9\%$ of memories receive no randomized exposure and none reaches the five-draw identification threshold, reducing discrimination from $0.765$ to $0.651$. Concentrating exposure on a candidate pool is therefore required by the arithmetic. This, however, requires a pool selector that contains the memories that matter. None of the deployable selectors in Appendix~\ref{app:baselines} provides adequate coverage, so pool selection remains the principal open problem.

\section{Reproducibility and Known Discrepancies}
\label{app:reproducibility}

\vspace{-2mm}
This section documents the evaluation instruments and implementation issues relevant to reproducing and interpreting the reported results.

\vspace{-2mm}
\subsection{Evaluation Instruments}
\label{app:repro_instruments}

\textbf{The LLM judge is not stable.} Rescoring the same $50$ responses $20$ times at temperature $0$ yields accuracies from $0.30$ to $0.38$. Across $2{,}000$ scoring calls, the provider reported 7 backend configurations, and variance estimates on identical inputs differed by up to a factor of three. Because this variability is comparable to the effects under study, we report only deterministic evaluation metrics.

\textbf{The judge scores abstention as success.} 
Across policies, $11\%$--$65\%$ of responses judged correct are refusals to answer. A policy that retrieves nothing and abstains is therefore rewarded, and correcting for this reverses the ranking of methods on LongMemEval. This is a property of the evaluation protocol rather than of any particular method.

\textbf{Exact match understates correctness on multi-hop data.} 
Gold answers in HotpotQA and MuSiQue sometimes include leading prepositions omitted by otherwise correct responses, causing exact match to score them as incorrect. We therefore report a containment metric alongside exact match and note whenever the two lead to different conclusions.

\textbf{Eight benchmark items are unanswerable as annotated.} 
Appendix~\ref{app:results_suspect} analyzes the eight \textsc{Suspect} items. Even with perfect retrieval, $44.6\%$ remain incorrect, establishing a performance ceiling inherited by any evaluation on this subset.

\subsection{Implementation Discrepancies}
\label{app:repro_implementation}

\textbf{The repository's default answer path leaks item metadata.} The serving code constructs a context line populated with the question type and question date, which reaches the generator. This does not affect any result in this paper: every experimental arm we report constructs its prompt from the retrieved memory texts and the question alone, verified by reconstructing the exact prompt from stored artifacts. But it means absolute scores obtained through the default path are not comparable to those reported here, and it is one reason the system-level comparison of Appendix~\ref{app:baseline_withheld} is withheld.

It also bears on Definition~\ref{def:scm}. In the arms we report, $c_t$ has no edge to $Y_t$ and retrieval mediates every path from the memory state to the output. Under the default path $c_t$ would be a direct cause of $Y_t$, though not a path from $M_t$, so Propositions~\ref{prop:factor} and~\ref{prop:nonid} are unaffected either way.

\textbf{Memory rendering differs across experiment families.} Some arms present a stored turn as raw content, others prefix a role or speaker label. All are deterministic functions of the retrieved set alone, so mediation and the identification results are unaffected, but absolute scores are not byte-comparable across families. Every comparison we report is within a family.

\textbf{Reasoning tokens cannot be disabled.} The generator consumes part of its output budget on reasoning tokens that are not returned. We log reasoning and visible tokens separately and flag every truncation. Truncation occurs on $2$ of $41{,}961$ calls ($0.005\%$) at a $1{,}024$-token budget, with mean output $190.8$ tokens and $p_{95} = 394$, and truncated cells are regenerated. A further $20$ calls ($0.048\%$) returned empty responses without truncating.

\subsection{Pool Construction Determines the Conclusion}
\label{app:repro_pool}

A methodological finding that our own measurements depend on. Evaluating selection policies over a small curated candidate set and over the full store can yield opposite conclusions on the same items with the same policies. On curated sets of at most ten candidates, recency-based selection performs within $0.03$ of an oracle; over full stores its recall of required memories is $0.000$ through a context budget of eight.  Similarity-based selection is significantly worse than random among topically similar candidates and significantly better among off-topic ones, with the sign reversing on pool size. 

We report every result in this paper over full stores for this reason, and we note it here because it is a property of how such comparisons are constructed.

\subsection{Compute and Cost}
\label{app:compute}

All experiments run on a single workstation with an $8$-core CPU and $64$\,GB of RAM; no GPU is required, as generation is served by API and embedding models run on CPU. Evaluation is parallelized across items with eight workers. Total API expenditure across the experiments reported here is \$$193.06$ over $80{,}187$ generation calls, of which the deployed-system replay accounts for \$$106.55$. Earlier exploratory runs on the main benchmarks were conducted without cost instrumentation and their expenditure is not recoverable, so this figure covers the experiments described in Appendices~\ref{app:protocols} and~\ref{app:additional_results}. Cost logs are written per experiment and are append-only; every reported figure is traceable to a logged call count and token count.

\subsection{Artifacts}
\label{app:artifacts}

We release the estimation code,  exposure assignment procedure,  evaluation harness, per-experiment protocols, and raw generated outputs with their scores. Every table in Section~\ref{sec:experiments} and in Appendix is reproducible from the released artifacts and recorded seeds. Assignment permutations, item selections, and pool constructions are seeded and the seeds are stored alongside the results they produced.

%% file: appendix/4_limitations.tex
\section{Limitations}
\vspace{-2mm}
%Three limitations bound what CMP establishes.

\textbf{Identification does not deliver a retention policy.}
CMP produces an identified measurement of per-query memory utility, and we show that this quantity does not by itself determine what to retain. A retention decision precedes the next query, so a policy must aggregate over queries, and every aggregation available to it is either attenuated by the fraction of queries for which a memory is required, or degenerate in the sense that it retains everything. The aggregation that would work uses information a deployed policy does not have. We therefore make no eviction claim, and closing this gap requires either a different estimand or a source of information about the query distribution that we do not assume.

\textbf{The exposure budget scales with the pool, not the store.}
By Corollary~\ref{cor:budget}, identifying the members of an exposure pool $\cP$ requires a horizon growing linearly in $|\cP|/k$. At realistic store sizes, uniform exposure leaves most memories with no randomized draws, so the design must concentrate exposure on a candidate set. That requires a pool selector containing the memories that matter, and the three deployable rules we test have near-zero nomination of required memories. Our demonstration uses a pool constructed with knowledge of which memories are required, which isolates the identification effect and does not establish that a practical selector recovers it. Exposure also has a direct cost while estimates are collected: displacing ranked slots changes answer quality by $-0.026$ F1 on average, not statistically distinguishable from zero but concentrated on queries the ranker already answers well.

\textbf{Scope.}
The identification failure binds on policies that rank memories by estimated utility. Heuristic policies are largely unaffected, not because they identify required memories but because required memories are accessed more often than typical ones, so frequency-, similarity-, and recency-based eviction routes around them. The result therefore constrains a class of methods rather than memory management in general. Our evidence is also drawn from a small number of benchmarks and one deployed system, and the level of the failure varies across them with the retrieval configuration.

%% file: appendix/5_applications.tex
\section{Applications of This Work}
\label{app:applications}

{\bf Retrieval-augmented generation and context attribution.} Retrieval-augmented generation conditions a model's output on documents retrieved from an external corpus~\citep{lewis2020retrieval}, and context attribution asks which of those documents caused the answer. The field has moved from generating citations toward measuring causal influence directly. ContextCite ablates context sources and fits a surrogate to the resulting changes in output~\citep{cohen2024contextcite}, Shapley-based methods value each retrieved document by its marginal contribution across subsets~\citep{nematov2025source}, and internals-based methods trace an answer to its context through the model's own activations~\citep{qi2024model}; the reliability of these attributions is now itself under study~\citep{abolghasemi2025evaluation}. Every one of these methods operates on documents already in the context. A document the retriever did not surface offers nothing to ablate, so attribution is silent about it and cannot distinguish a document that is irrelevant from one that was never tested. This is Assumption~\ref{assum:ret_positivity} failing in a second setting. Attribution explains the answer the system gave, but it cannot say whether a better answer was available from evidence the retriever passed over. Proposition~\ref{prop:factor} separates the two quantities attribution conflates, retrievability and conditional usefulness, and randomized exposure supplies the arm attribution lacks, so the contribution of an unretrieved document becomes estimable at the cost of displacing some ranked evidence while estimates are collected.

{\bf Data valuation and content compensation.} Data valuation assigns each data source a measure of its contribution to a model's behavior, and as generative search replaces ranked links with synthesized answers, it is increasingly proposed as the basis for compensating the providers of that content. Data Shapley established valuation for training data~\citep{ghorbani2019data}. Recent work extends it to documents supplied as context at inference time, valuing the sources of LLM-generated summaries with clustered Shapley approximations~\citep{ye2025document} and adapting payment rules from mechanism design to retrieval-augmented data markets~\citep{fan2025data}. These methods value the documents a retriever surfaces. Under leave-one-out and mechanism-design payments only retrieved documents are compensated~\citep{fan2025data}, so a document the ranker never places in context receives nothing regardless of what it would have contributed. The undervaluation is systematic: it tracks retrieval rank, and a document that is never retrieved generates no evidence that it should be. Theorem~\ref{prop:nonid} explains why collecting more logs under the same retriever cannot correct this. CMP's exposure design gives every document in a pool a known, positive probability of entering the context, which separates a document's value from the ranker's assessment of it and makes the value of unretrieved sources identifiable.

{\bf Recommender systems and counterfactual evaluation.} Recommender systems learn from logged feedback that exists only for items the serving policy chose to show, which makes exposure bias a central problem in the field. Inverse-propensity methods correct for it in learning to rank~\citep{joachims2017unbiased}, feedback loops in which unexposed items remain unexposed are well documented~\citep{jiang2019degenerate}, and two-stage architectures, in which a candidate-generation stage gates what a ranking stage can evaluate, have required off-policy corrections of their own~\citep{ma2020off}. The problem has now reached LLM-based recommenders, where self-normalized inverse propensity scoring corrects exposure bias in continual post-deployment updates~\citep{kim2026don}. This literature largely treats an unexposed item as recoverable: an item not shown today can be explored tomorrow, so exposure bias is a problem of estimation efficiency. That assumption fails whenever an action removes an item from future support, as in catalog delisting, inventory retirement, or memory eviction, where a positivity violation becomes permanent. Two of our results carry over. The mediation of Section~\ref{sec:scm} is the two-stage architecture, and identifying an item the candidate stage never passes requires randomizing that stage itself, as Section~\ref{sec:aipw} does for retrieval. Theorem~\ref{prop:abstention} characterizes the Bayes decision when an action destroys the ability to re-estimate, which is the decision a platform faces when it retires items on the basis of exposure-biased logs.

{\bf Cognitive science of memory.} Cognitive psychology has long distinguished whether a memory is stored from whether it can be retrieved. Items a person fails to recall are often still available and can be recovered with an appropriate cue~\citep{tulving1966availability}, a dependence encoding specificity attributes to the match between the retrieval cue and the conditions of storage~\citep{tulving1973encoding}. The theory of disuse formalizes the two as separate quantities, storage strength and retrieval strength~\citep{bjork1992new}, and retrieval is competitive, since retrieving some items suppresses related ones~\citep{anderson1994remembering}. Complementary learning systems theory has carried related ideas into models of consolidation~\citep{kumaran2016learning, mcclelland2020integration}, but the distinction between storage and access has not informed how artificial memory is valued. Memory systems infer a memory's value from behavior, which conflates the two quantities cognitive science keeps apart. Proposition~\ref{prop:factor} is their computational form: presence in the store is availability, $P(m \in R \mid q, c)$ is accessibility, and $U_R$ is the value of a memory once accessed. The parallel extends to method. Cognitive science revealed available but inaccessible memories by changing the cue, and randomized exposure plays the same role, a cued-recall manipulation run with known probability. Retrieval competition has an analogue as well: a deterministic ranker that repeatedly surfaces the same memories deprives the rest of evidence, and utility-ranked eviction then removes them.

{\bf Human--AI interaction and longitudinal relationships.} Persistent memory turns interaction with an assistant into a relationship that accumulates across sessions. Commercial assistants now retain information about users between conversations and offer controls to inspect and delete it, and interface research has made stored memories visible and editable~\citep{huang2023memory}. A recent study of ChatGPT's memory feature found that users perceive AI memory as unforgetful, that most experienced violated expectations on learning what it retained, and that they want greater visibility and control~\citep{chen2026relational}. Appropriate reliance on such a system requires that users can tell when it errs~\citep{schemmer2023appropriate}. Longitudinal clinical records raise the same question in a high-stakes form: causal explanation with graph neural networks has been applied to the progression of mild cognitive impairment~\citep{behnam2024causal}, and a memory-augmented assistant serving such a user faces costly irreversible errors whenever it forgets. Transparency interfaces show what a memory system stores, not what it uses, and on LongMemEval the two differ substantially: $54\%$ of required memories are never retrieved. Such a memory appears in a memory viewer yet is functionally forgotten, so the user's belief that the assistant does not forget diverges from its behavior, and the error is unobservable, since the answer the memory would have produced is never shown. The divergence compounds over longitudinal use. A memory that is never retrieved produces no evidence of its value, utility-ranked eviction then deletes it, and information the user chose to share disappears without the system having tested whether it mattered. Proposition~\ref{prop:factor} distinguishes a stored memory from an accessible one, the distinction an honest memory interface would display, and randomized exposure lets a system learn whether a shared memory helps.

{\bf Evaluating long-term memory in LLM agents.} The application closest to this work is the evaluation of agent memory itself. LongMemEval~\citep{wu2024longmemeval} and LoCoMo~\citep{maharana2024evaluating} established long-horizon recall as the standard test, and a second generation of benchmarks extends it to memory that changes over time: whether agents recognize that a memory is no longer valid~\citep{chao2605stale}, how they handle forgetting and test-time learning~\citep{hu2026evaluating}, and how memory supports agents in environments beyond dialogue~\citep{di2605longmemeval}. All of these query a system and score its answer, which measures recall for the questions asked. A retention decision depends on something else, a memory's value for questions not yet asked, and a benchmark that queries each store once cannot measure it. LongMemEval does exactly that: every store carries a single question, and among its multi-evidence items none of the $203$ required memories is needed by a second question. Our framework specifies what retention evaluation requires, namely several queries against one store with overlapping required evidence, and LoCoMo meets it, with $750$ of $1{,}425$ required memories shared across questions. It also supplies a validity check that can be run before any policy is compared: a benchmark whose required memories the evaluated retriever never surfaces measures the retriever, and Assumption~\ref{assum:ret_positivity} can be tested on it directly.